\documentclass{article} 
\usepackage{iclr2026_conference,times}
\renewcommand{\headrulewidth}{0pt}

\usepackage{amsmath,amsfonts,bm}

\def\eqref#1{equation~\ref{#1}}

\def\1{\bm{1}}

\DeclareMathAlphabet{\mathsfit}{\encodingdefault}{\sfdefault}{m}{sl}
\SetMathAlphabet{\mathsfit}{bold}{\encodingdefault}{\sfdefault}{bx}{n}

\newcommand{\R}{\mathbb{R}}

\usepackage[colorlinks=true]{hyperref}
\hypersetup{
    colorlinks,
    citecolor=[rgb]{0.1, 0.3, 0.4},
    linkcolor=[rgb]{0.1, 0.3, 0.4},
    urlcolor=[rgb]{0.1, 0.3, 0.4},
}
\usepackage{url}
\usepackage{cleveref}

\usepackage{amssymb}
\usepackage{pifont}
\usepackage{mathtools}
\usepackage{algorithm}
\usepackage{algpseudocode}
\usepackage{bbm}
\usepackage{csquotes}
\usepackage{graphicx}
\usepackage{subcaption}
\usepackage{booktabs}
\usepackage{wrapfig}
\usepackage{comment}
\usepackage{amsthm}
\usepackage{enumitem}
\usepackage{xcolor}
\definecolor{terminalbg}{HTML}{1E1E2E}
\usepackage{tcolorbox}
\usepackage{listings}
\usepackage{placeins}
\usepackage{xspace}
\usepackage{tikz}
\usetikzlibrary{positioning, arrows.meta, decorations.pathreplacing, calc, fit, backgrounds}
\usepackage{tabularx}
\usepackage{multirow}

\newtcolorbox{commandmentbox}[1][]{%
    colback=gray!5!white,
    colframe=gray!85!black,
    fonttitle=\bfseries,
    title=#1,
    boxrule=0.5pt,
    arc=2pt,
    left=6pt,
    right=6pt,
    top=4pt,
    bottom=4pt,
}

\newcommand{\cmdref}[1]{\hyperref[cmd:#1]{Commandment~#1}}

\lstdefinestyle{claudemd}{%
    basicstyle=\ttfamily\small,
    backgroundcolor=\color{gray!8},
    frame=single,
    framerule=0.4pt,
    rulecolor=\color{gray!40},
    breaklines=true,
    columns=fullflexible,
    keepspaces=true,
    xleftmargin=4pt,
    xrightmargin=4pt,
    aboveskip=6pt,
    belowskip=6pt,
}

\newcounter{casestudy}

\crefname{casestudy}{Case Study}{Case Studies}
\Crefname{casestudy}{Case Study}{Case Studies}

\newcommand{\giturl}{\href{https://github.com/ZIB-IOL/The-Agentic-Researcher}{github.com/ZIB-IOL/The-Agentic-Researcher}\xspace}

\title{\centering{The Agentic Researcher:} \\A Practical Guide to AI-Assisted Research in Mathematics and Machine Learning}

\author{Max Zimmer\thanks{We welcome contributions, issue reports, improvement suggestions, additional case studies via issues, PR, \giturl, to keep this up-to-date and useful.} \quad Nico Pelleriti \quad Christophe Roux \quad Sebastian Pokutta\\
Department for AI in Society, Science, and Technology, Zuse Institute Berlin, Germany\\
Institute of Mathematics, Technische Universit\"at Berlin, Germany\\
\texttt{\{zimmer, pelleriti, roux, pokutta\}@zib.de} \\
}

\iclrfinalcopy
\begin{document}

\maketitle
\vspace{-18pt}
\begin{figure}[h]
  \begin{minipage}{\linewidth}
  \begin{tcolorbox}[
    colback=terminalbg, colframe=terminalbg,
    arc=6pt, boxrule=0.5pt,
    left=3pt, right=3pt, top=3pt, bottom=3pt,
    boxsep=0pt
  ]
  \begin{center}
  \includegraphics[width=.85\linewidth]{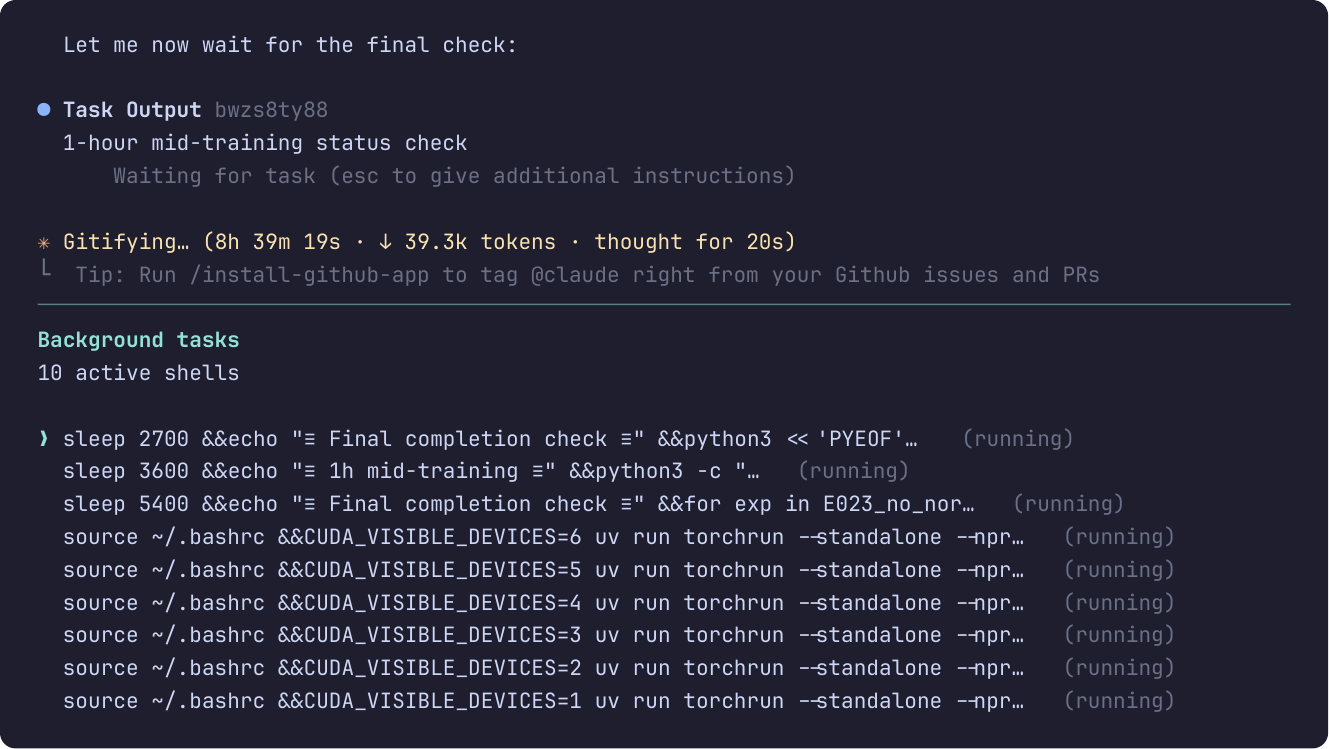}
  \end{center}
  \end{tcolorbox}
  \captionof{figure}{A command-line interface (CLI) agent during an autonomous research session: over 8 hours in, managing six parallel GPU training runs and three scheduled monitoring tasks. The same framework supports \emph{mathematical derivations, proofs, and verification} alongside computational experiments. The agent is idle, consuming no tokens while waiting for a status check to complete.}
  \label{fig:terminal_session}
  \end{minipage}
  \end{figure}
 
\begin{abstract}
  \vspace{-10pt}
  AI tools and agents are reshaping how researchers work, from proving theorems to training neural networks. Yet for many, it remains unclear how these tools fit into everyday research practice.
  This paper is a \emph{practical guide to AI-assisted research in mathematics and machine learning}: We discuss how researchers can use modern AI systems productively, where these systems help most, and what kinds of guardrails are needed to use them responsibly.
  It is organized into three parts: (I)~a five-level taxonomy of AI integration, (II)~an open-source framework that, through a set of methodological rules formulated as agent prompts, turns CLI coding agents (e.g., Claude Code, Codex CLI, OpenCode) into autonomous research assistants, and (III)~case studies from deep learning and mathematics.
  The framework runs inside a sandboxed container, works with any frontier LLM through existing CLI agents, is simple enough to install and use within minutes, and scales from personal-laptop prototyping to multi-node, multi-GPU experimentation across compute clusters. In practice, our longest autonomous session ran for over 20 hours, dispatching independent experiments across multiple nodes without human intervention.
  We stress that our framework is not intended to replace the researcher in the loop, but to augment them.
  Our code is publicly available at \giturl.
  \end{abstract}

\section{Introduction}
\label{sec:introduction}

In 2024, DeepMind's AlphaProof~\citep{hubertOlympiadlevelFormalMathematical2025} combined with AlphaGeometry~\citep{trinhSolvingOlympiadGeometry2024a} became the first AI system to achieve medal-level performance at the International Mathematical Olympiad (IMO), reaching silver-medal standard by solving four of the six competition problems through reinforcement learning and formal verification.
AlphaEvolve~\citep{novikovAlphaEvolveCodingAgent2025} demonstrated that LLM-guided evolutionary search can discover new mathematical constructions, rediscovering best-known solutions across a broad collection of problems and improving on them in several cases~\citep{georgievMathematicalExplorationDiscovery2025}.
Most recently, Aletheia~\citep{fengAutonomousMathematicsResearch2026}, an autonomous mathematical research agent, resolved several open problems originally posed by Erd\H{o}s while operating with minimal human intervention. Aletheia also solved several open problems from \emph{First Proof}~\citep{abouzaid2026}, a benchmark of previously unpublished research-level mathematics questions drawn from the authors' own research process, within \emph{weeks} of its release. These results are remarkable, and recent systems now address not only well-defined benchmarks but also genuine open mathematical problems. In parallel, the Machine Learning (ML) community has seen a surge in agentic experimentation: for instance, Karpathy's \emph{autoresearch}~\citep{karpathy_autoresearch_2026} demonstrated how agents can run automated ML experiment pipelines through iterative code modification, and such pipelines are becoming increasingly common.

Most of the current literature, including the works discussed above, focuses on \emph{what AI systems can achieve}. Much less attention has been given to the complementary practical question of \emph{how researchers should integrate such systems} into everyday research.
In practice, research rarely proceeds by pursuing a fixed objective from the outset: researchers must decide which questions to ask, which experiments to run, when to reformulate a conjecture, and how to respond to unexpected results.
Supporting this kind of work requires workflows that accommodate shifting objectives, iterative experimentation, and sustained human guidance, yet how to build and use such workflows remains an open question.
For most researchers, the challenge is not building a discovery pipeline from scratch but understanding which tools are available and how to use them effectively.

A growing body of work has begun to map this landscape, including conceptual frameworks for human-AI co-creativity~\citep{haaseHumanAICoCreativityExploring2026}, visions of the ``augmented mathematician''~\citep{henkelMathematiciansAssistantIntegrating2025}, formal-proof assistants~\citep{yangLeanDojoTheoremProving2023, songLeanCopilotLarge2025}, and numerous first-hand accounts of AI-assisted research~\citep{bubeck2025a, diez2025a, alexeev2026, ivanisvili2025a, feldman2025a, salim2025a, dobribanSolvingResearchProblem2025, schmitt2025}.
\citet{avigadMathematiciansAgeAI} make this point especially clearly: mathematicians should not merely react to AI but should take an active role in deploying and shaping it for their own purposes.
Yet none of these works provides actionable, end-to-end guidance that a researcher could follow today.

We hope to make \emph{some} progress on these questions and aim to fill \emph{parts of} that gap. The frameworks, approaches, and insights presented here have been developed over roughly the last one and a half years in the context of the MATH+ project \emph{Agentic AI in Mathematics}\footnote{\url{https://iol.zib.de/project/agentmath.html}} but apply beyond mathematics and have proven to be very powerful, e.g., in ML research. This also explains our choice of use cases in machine learning and mathematics.
The four authors approached AI-assisted research from complementary directions: some built on existing CLI coding agents with either an experimental or a theoretical and proof-oriented focus, while others developed a custom multi-agent system from scratch. The insights gained from these diverse experiences form the basis of the unified framework we present here.

\paragraph{Contributions.} Our contributions are as follows.

\begin{enumerate}[leftmargin=*, itemsep=2pt]
    \item \textbf{A practical taxonomy} (\Cref{sec:levels}). We identify five levels of AI integration into mathematical and ML research, ranging from full human control to high agent autonomy.

    \item \textbf{An open-source, sandboxed agentic research framework} (\Cref{sec:framework}). We present a set of methodological rules, formulated as agent prompts, which we call \emph{commandments}, together with a sandboxed container environment and reporting conventions that turn general-purpose CLI coding agents into autonomous research assistants. The commandments encode the norms of scientific practice and guide the agent throughout the research workflow. The framework is model- and harness-agnostic, supports any frontier LLM through existing CLI agents (such as Claude Code~\citep{ClaudeCodeOverview}, Codex CLI~\citep{CodexAICoding}, or OpenCode~\citep{AnomalycoOpencode2026}), and can be set up within minutes.

    \item \textbf{Case studies} (\Cref{sec:case_studies}). We demonstrate the framework in action across diverse domains, including deep learning as well as pure and applied mathematics, illustrating both successes and failure modes. We provide screenshots of the agent's reports as they were produced.
\end{enumerate}

We want to emphasize what this paper is \emph{not}: we do not claim that AI replaces research creativity, insight, or the researcher. Rather, we demonstrate that specific parts of the research workflow can be significantly accelerated when a researcher directs an AI agent in a structured way. Unlike approaches that seemingly remove the human from the research process entirely \citep[cf., e.g.,][]{luAIScientistFully2024}, our framework keeps the researcher as the principal investigator, who can now operate at greater scale and speed. We believe that mathematical research is not a fully automatable task, and we will not speculate on whether this will change in the future. What we do claim is that mathematicians and researchers in general should take an active role in this partial transformation of the field and, echoing \citet{avigadMathematiciansAgeAI}, should \emph{own} the technology.

The rest of this paper is organized as follows.
\Cref{sec:levels} presents our taxonomy of integration levels.
\Cref{sec:framework} describes the agentic research framework in detail, the core contribution of this paper.
\Cref{sec:case_studies} presents case studies, and \Cref{sec:discussion} concludes with lessons learned, limitations, and future directions. We defer the survey of related work to \Cref{sec:related_work} at the end of the paper.

\section{Levels of AI Integration in Mathematical and ML Research}
\label{sec:levels}
Inspired by \citet{haaseHumanAICoCreativityExploring2026}, we propose a taxonomy of five levels that characterize how deeply AI is integrated into the research process, ranging from no AI involvement to fully autonomous research loops.
These levels are not mutually exclusive, and a researcher might use different levels for different tasks, all within the same project. In particular, even (fully) autonomous systems can delegate subtasks to less autonomous components. This regularly happens also in our setup when subagents are spawned to accomplish subtasks. In general, the key lies in recognizing which level is appropriate for which task. \Cref{tab:levels} summarizes the taxonomy, and we describe each level in detail below.

\begin{table}[b]
  \centering
  \caption{Five levels of AI integration in mathematical research. Each (not necessarily mutually exclusive) level represents a qualitatively different trade-off between agent autonomy and human involvement.}
  \label{tab:levels}
  {
  \setlength{\tabcolsep}{3pt}
  \renewcommand{\arraystretch}{1.3}
  \footnotesize
  \begin{tabularx}{\textwidth}{c l >{\raggedright\arraybackslash}p{2.6cm} X >{\raggedright\arraybackslash}p{2.9cm}}
  \toprule
  \textbf{Level} & \textbf{Name} & \textbf{Tools} & \textbf{AI Tasks} & \textbf{Human Role} \\
  \midrule
  0 & Classical & \LaTeX{}, math.\ software & No AI integration & Everything \\
  1 & Consultant & LLM chatbots & Targeted queries for explanation, literature, brainstorming & Asks, evaluates \\
  2 & Typist & Editor plugins (Copilot, Cursor) & Code and text generation without execution & Thinks, reviews, decides \\
  3 & Collaborator & CLI coding agents & Human describes task, AI implements and iterates & Reviews each output, assigns next task \\
  4 & Research Assoc. & Our framework & Autonomous experiment loop following structured research plan & Steers, audits \\
  \bottomrule
  \end{tabularx}
  }
  \end{table}

\paragraph{Level 0: Classical.}
The classical level is the baseline of our taxonomy and the traditional mode of mathematical and ML research.
The researcher uses all traditional computational tools, including typesetting software (e.g., \LaTeX{}), mathematical software (e.g., Mathematica, MATLAB), and programming languages for custom implementations (e.g., Python, Julia, PyTorch), but no AI assistance.
This remains the predominant mode of research and is perfectly appropriate. The goal of this paper is not to argue that AI should render it obsolete, but to show when and how AI can complement it.

\paragraph{Level 1: AI as Consultant.}
The researcher uses LLM-based chatbots (e.g., ChatGPT, Claude, Gemini) for specific queries and assistance. Typical cases include concept explanation (\emph{Explain the difference between strong and weak duality in linear programming}), literature search (\emph{What are the current best convergence rates for SGD with heavy-tailed noise?}), brainstorming (\emph{What techniques exist for proving convergence of iterative algorithms when the operator is only approximately contractive?}), and debugging ideas (\emph{Here is my proof attempt. Where does the argument break down?}).

The core intellectual work remains with the researcher; the AI provides targeted assistance.
The key skill is asking the right questions and crafting sufficiently detailed prompts to guide the AI toward a useful answer.
A clear limitation is that the interaction is stateless across sessions unless the user manually provides context.

\noindent\emph{Getting started:} A web browser and access to an LLM chatbot (free tiers available from most providers). No setup required.

\paragraph{Level 2: AI as Typist.}
The researcher uses AI for code and text generation, ranging from tab completion (e.g., GitHub Copilot predicting the next line) to more complex prompt-based generation that produces entire functions or \LaTeX{} paragraphs from a natural-language description.
Every output is reviewed by the researcher and accepted, edited, or rejected.

The defining characteristic of this level is that the AI generates code or text but neither executes nor iterates on the results.
The researcher remains responsible for all design decisions, and the AI accelerates the writing process without closing the loop between implementation and evaluation.

\noindent\emph{Getting started:} Install a code editor plugin (e.g., Cursor, or VS Code with GitHub Copilot).

\paragraph{Level 3: AI as Collaborator.}
The full implementation and execution are delegated to a \emph{CLI coding agent}, i.e., a terminal-based tool (e.g., Claude Code~\citep{ClaudeCodeOverview}, OpenCode~\citep{AnomalycoOpencode2026}, Codex CLI~\citep{CodexAICoding}) that can read and edit files, execute shell commands, and iterate on results within a persistent project context.
This differs qualitatively from Levels~1--2 because the agent possesses a much broader set of capabilities, including file modifications, code execution, and iteration based on results it has obtained, all within a single conversation.
For a prompt like \emph{``Implement the Frank-Wolfe algorithm for the semidefinite relaxation of max-cut, with step size $\gamma_t = 2/(t+2)$''} or \emph{``Implement a learning rate scheduler with linear warmup,''} the agent reads the codebase, implements the algorithm, runs it, and re-evaluates if convergence shows unexpected behavior.

The researcher describes each task in natural language and provides the necessary context, such as an existing codebase.
After each completed task, the researcher reviews the output, decides \emph{what} to do next, and assigns the next task; the agent handles \emph{how}.
At no point does the agent independently set the research direction.

\noindent\emph{Getting started:} Install a CLI coding agent and start a session in the project directory.

\paragraph{Level 4: AI as Research Associate.}
The highest degree of autonomy in our taxonomy.
The researcher arrives with a research idea (initial intuitions, failed strategies, partial results, or simply a well-posed question) and outlines a research plan: goals, metrics, constraints, approaches already tried, and promising directions to explore.
The agent then formulates a detailed plan and autonomously executes an experiment loop: formalizing mathematical ideas, implementing approaches, running evaluations, recording results, analyzing outcomes, and updating both a structured research report and a \texttt{TODO.md}. It iterates this loop, continuously refining and expanding the plan, operating for hours to days to achieve the research goal or uncover something unexpected.

To operate for extended periods, structured and clear instructions that govern scientific rigor, documentation, and verification are needed: our framework (\Cref{sec:framework}) provides exactly these.
The key difference from Level~3 is that the agent does not wait for human input between experiments but follows a research plan and a set of commandments encoding the norms of good scientific practice: one variable per experiment, structured reporting, staged evaluation (from quick sanity checks to full benchmarks), and verification protocols, among others (cf.\ \Cref{sec:framework}).
Intermittent human review and course correction are an integral part of Level~4, not a fallback to Level~3: the researcher periodically inspects the report, adjusts priorities, and refines the research plan while the agent continues to execute autonomously.
The researcher's role shifts from execution to direction-setting, periodic review, and evaluation.
Level~4 is most appropriate when the search space is large.

Despite the guardrails described in \Cref{sec:framework}, limitations remain.
The agent may pursue an unproductive direction for too long, especially when the research plan lacks sufficient detail.
Verification is only partially solved: while we provide strategies for symbolic and numerical verification of mathematical claims and implementations, a high (to full) degree of certainty requires the researcher to perform a rigorous review of the work. We consider this a feature, not a bug.
Similarly, while the agent is instructed to search the literature, it cannot guarantee that its ideas are genuinely novel. Thorough knowledge of the related work remains the researcher's responsibility.
As such, the researcher still faces a non-trivial amount of work both throughout and toward the end of a project: reviewing intermediate results and providing steering, verifying correctness, deciding what results merit publication, and confirming originality as well as adding context and interpretation.
However, instead of conducting the entire research process alone, the researcher now externalizes parts of the work to a capable \emph{research associate} who delivers a structured, well-documented report. This report then requires careful and rigorous review with subsequent steering and guidance. Through repeated interactions of this kind, new results emerge in a process of Human-AI co-creation.

\noindent\emph{Getting started:} Clone the project repository\footnote{\giturl} and follow the setup instructions in the \texttt{README.md}. The setup takes a couple of minutes, and the first autonomous experiments can begin immediately. A detailed description of the framework initialization is given in \Cref{sec:setup}.

\begin{figure}[t]
  \centering
  \resizebox{\textwidth}{!}{
\begin{tikzpicture}[
    >=Stealth,
    cheader/.style={rectangle, rounded corners=3pt,
                    draw=orange!70!black, thick, fill=orange!15,
                    minimum height=0.85cm, minimum width=4.6cm, align=center,
                    font=\small\bfseries},
    pcontent/.style={rectangle, rounded corners=3pt,
                     draw=orange!70!black, thick, fill=orange!5,
                     minimum height=1.5cm, minimum width=4.6cm, align=center,
                     font=\footnotesize, text=black!80},
    exnode/.style={rectangle, rounded corners=2pt, draw=black!25, fill=gray!4,
                   minimum height=1.3cm, minimum width=4.6cm, align=center,
                   font=\footnotesize, text=black!75},
    bandlabel/.style={font=\footnotesize\bfseries, anchor=east, text=black!50},
    domainlabel/.style={font=\footnotesize\bfseries\scshape, anchor=east},
    exlabel/.style={font=\footnotesize, anchor=east, text=black!55},
]

\def\colA{2.0}
\def\colB{7.8}
\def\colC{13.6}
\def\labelX{-0.6}

\def\rowHead{0.9}
\def\rowExA{-2.9}
\def\rowExD{-4.7}

\node[bandlabel, text=orange!60!black] at (\labelX, \rowHead) {Concept};

\node[cheader] (h1) at (\colA, \rowHead) {Research Question};
\node[pcontent, anchor=north] (p1) at ([yshift=-3pt]h1.south) {%
    Problem formulation,\\[-1pt]
    hypotheses, objectives,\\[-1pt]
    evaluation criteria};

\node[cheader, draw=yellow!70!black, thick, fill=yellow!25] (h2) at (\colB, \rowHead) {Tools, Methods \& Data};
\node[pcontent, draw=yellow!70!black, thick, fill=yellow!12, text=black, anchor=north] (p2) at ([yshift=-3pt]h2.south) {%
    Software stack \& packages,\\[-1pt]
    datasets, compute resources,\\[-1pt]
    custom scripts};

\node[cheader] (h3) at (\colC, \rowHead) {Prior Work \&\\[-2pt] Domain Knowledge};
\node[pcontent, anchor=north] (p3) at ([yshift=-3pt]h3.south) {%
    Existing codebase,\\[-1pt]
    \LaTeX{} notes \& derivations,\\[-1pt]
    references, preliminary results};

\path let \p1=(p1.center) in node[bandlabel, text=orange!55!black] at (\labelX, \y1) {In practice};

\draw[black!20, thick] (current bounding box.west |- 0,-1.7) -- (current bounding box.east |- 0,-1.7);
\node[font=\footnotesize\bfseries, text=black!40, anchor=west] at (current bounding box.west |- 0,-2.05) {Examples:};

\node[domainlabel, text=orange!60!black] at (\labelX, \rowExA+0.2) {Case Study~A};
\node[exlabel, text=orange!50!black] at (\labelX, \rowExA-0.1) {\emph{Deep Learning}};
\node[exnode, draw=orange!70!black, thick, fill=orange!4] at (\colA, \rowExA) {%
    Improve LLM pretraining:\\[-1pt]
    exploit Muon's memory\\[-1pt]
    savings over AdamW};
\node[exnode, draw=yellow!70!black, thick, fill=yellow!10, text=black] at (\colB, \rowExA) {%
    PyTorch, CUDA, \texttt{uv},\\[-1pt]
    FineWeb dataset,\\[-1pt]
    multi-GPU allocation};
\node[exnode, draw=orange!70!black, thick, fill=orange!4] at (\colC, \rowExA) {%
    LLM pretraining\\[-1pt]
    benchmark codebase};

\node[domainlabel, text=orange!60!black] at (\labelX, \rowExD+0.2) {Case Study~D};
\node[exlabel, text=orange!50!black] at (\labelX, \rowExD-0.1) {\emph{Mathematics}};
\node[exnode, draw=orange!70!black, thick, fill=orange!4] at (\colA, \rowExD) {%
    Prove lower bounds for\\[-1pt]
    Frank-Wolfe on uniformly\\[-1pt]
    convex sets};
\node[exnode, draw=yellow!70!black, thick, fill=yellow!10, text=black] at (\colB, \rowExD) {%
    Python, Julia, \texttt{uv}};
\node[exnode, draw=orange!70!black, thick, fill=orange!4] at (\colC, \rowExD) {%
    Two recent lower-bound\\[-1pt]
    proofs for the strongly\\[-1pt]
    convex case as references};

\end{tikzpicture}}
  \caption{Setting up a research project.
  \textbf{Top:} the three categories of input the researcher provides, with their conceptual role (dark) and concrete realization (light).
  \textbf{Bottom:} two examples from our case studies: a deep learning project (\Cref{sec:case_a}) and a mathematics project (\Cref{sec:case_d}).}
  \label{fig:two_layers}
\end{figure}
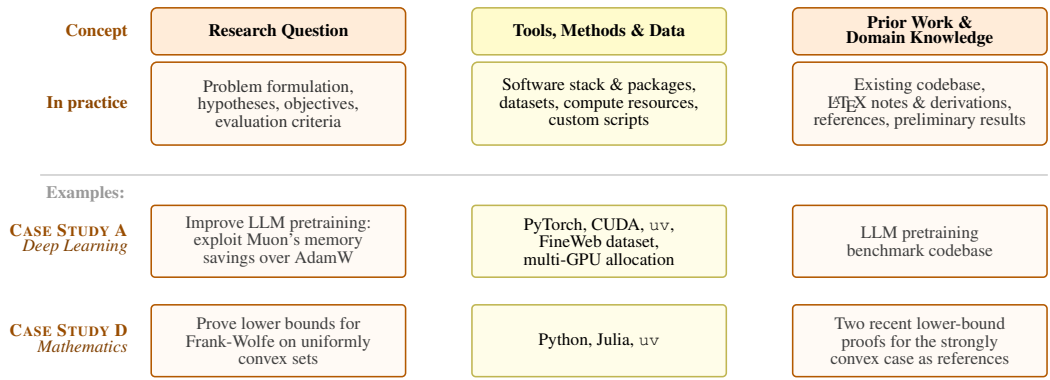

\section{The Agentic Research Framework}
\label{sec:framework}

We describe our core contribution: the agentic research framework, its design principles, and the ten commandments, distilled from our own experience, that guide the agent's behavior. The instructions described in the following subsections are provided to the agent through a persistent instruction file (\texttt{INSTRUCTIONS.md}) that is read at the start of every session. This configuration file contains universal instructions as well as a final section that serves as a template placeholder for project-specific instructions; these are automatically filled in by the agent once the researcher provides the research instructions.

\subsection{Overview and Workflow}
\label{sec:setup}

To start a new project, the researcher provides three things (\Cref{fig:two_layers}): a \emph{research question} (problem formulation, hypotheses, evaluation criteria), the \emph{tools, methods, and data} needed to investigate it (software stack, packages, datasets, compute resources), and any \emph{prior work or domain knowledge} that should inform the investigation (existing codebase, \LaTeX{} notes with derivations, references, preliminary results). In the following, we will use the term \emph{experiment} to refer to one (broad) agentic iteration loop with the researcher: depending on the context, this can be one proof attempt, an actual computational experiment, or the design of a new algorithm. The framework is built around CLI coding agents, e.g., Claude Code~\citep{ClaudeCodeOverview}, Codex CLI~\citep{CodexAICoding}, Gemini CLI~\citep{BuildDebugDeployGeminiCLI}, or OpenCode~\citep{AnomalycoOpencode2026}, which operate inside a sandboxed container that provides a secure, isolated workspace.

\paragraph{Starting a new project.}
The typical workflow is as follows:
\begin{enumerate}[leftmargin=*, itemsep=1pt]
  \item The researcher begins in a project directory that contains the practical-layer materials described above (\Cref{fig:two_layers}).
  From this directory, they launch the sandbox and provide the research instructions to the agent. The more detailed the instructions, the better; we found it especially useful to provide a working codebase if one exists, along with a \LaTeX{} write-up of the research problem and previously tried approaches.
  \item The agent asks clarifying questions about scope, constraints, and evaluation metrics.
  \item After this back-and-forth, the agent explores all relevant files and writes the final project-specific instructions into a persistent instruction file (\texttt{INSTRUCTIONS.md}), alongside the universal commandments that are already in place (\Cref{sec:commandments}).
  \item The agent creates a plan and initializes \texttt{report.tex} and \texttt{TODO.md}, the two main artifacts of the research process. Upon approval by the researcher (or after further refinement of the plan), the agent begins autonomous execution and only requires human intervention in case of unexpected behavior or when the research plan needs adjustment.
\end{enumerate}

\paragraph{Why CLI agents.}
Across our research workflows, three practical requirements arose repeatedly.
CLI agents are \emph{easy to use}: they fit naturally into local working environments, can be launched inside an existing project, and operate directly on local files without additional infrastructure.
They remain fully \emph{interactive}: the researcher can intervene at any point to inspect progress, redirect the investigation, stop execution, or restart with revised instructions.
Finally, they are \emph{extensible}: the toolchain can be readily extended with custom utilities; in our case, this included scripts for handling literature and \LaTeX{} sources, extracting relevant algorithmic sections, and running specialized search and verification routines. The same mechanism also supports hard \emph{guardrails}: automated checks can be triggered after file edits or experiment runs, enforcing formatting, running tests, or updating reports. Because CLI agents are maintained by model providers and evolve with model capabilities, while our rules sit on top, the framework automatically benefits from improvements to the underlying tools. \Cref{fig:terminal_session} shows an autonomous session in practice.

\paragraph{Infrastructure.}
Because the framework is built around CLI agents, the surrounding infrastructure can remain intentionally minimal.
The sandbox confines all actions to a container, enabling unattended sessions without the risk of damaging the host system.
For compute-intensive projects, a multi-node launcher dispatches independent experiments to remote Slurm nodes.
We recommend using reproducible, project-local package managers (\texttt{uv} for Python, Julia's \texttt{Pkg}, among others).

\paragraph{Structured reporting and experiment tracking.}
All experimental progress is recorded in a single \LaTeX{} file (\texttt{report.tex}) that accumulates experiments, derivations, and analysis, complemented by a \texttt{TODO.md} checklist for open questions, unverified claims, and deferred work.
Each experiment subsection must contain the following fields, enforced by the commandments (\Cref{sec:commandments}):

\begin{minipage}{\linewidth}
\begin{lstlisting}[style=claudemd, caption={Required fields for each experiment in \texttt{report.tex}.}]
\paragraph{Goal} What problem are we solving?
\paragraph{Hypothesis} Why should this approach work?
\paragraph{Method} Mathematical formulation with proper notation.
\paragraph{Implementation} Files and lines changed.
\paragraph{Results} Table with method, model/instance, metric, delta.
\paragraph{Analysis} Why it worked or didn't. What it reveals.
\paragraph{Next Steps} What to try based on these results.
\end{lstlisting}
\end{minipage}

\noindent Rather than introducing a separate experiment-tracking system, we use Git directly.
Each experiment is recorded as a commit with a structured message of the form \texttt{exp(EXXX): <description> -- <metric>=<value>}.
Branches group related experiments, tags mark important outcomes, and Git's worktree feature allows multiple agent sessions to run concurrently on separate copies of the codebase without interference.
This keeps the full experimental history lightweight, portable, and directly searchable through Git logs.

Once running, each experiment follows the eight-step loop shown in \Cref{fig:framework}: \emph{Explore}~$\to$ \emph{Plan}~$\to$ \emph{Implement}~$\to$ \emph{Evaluate}~$\to$ \emph{Analyze}~$\to$ \emph{Record}~$\to$ \emph{Commit}~$\to$ \emph{Iterate}. At the beginning of every session (or after a context window reset), the agent re-reads \texttt{report.tex}, \texttt{TODO.md}, and the \texttt{git log} to restore continuity.

\begin{figure}
  \centering
  \resizebox{\textwidth}{!}{
\resizebox{\textwidth}{!}{%
\begin{tikzpicture}[
    >=Stealth,
    actor/.style={rectangle, rounded corners=3pt, draw=black!70, fill=blue!6,
                  minimum height=0.85cm, minimum width=2.6cm, align=center,
                  font=\small\bfseries},
    component/.style={rectangle, rounded corners=2pt, draw=black!50, fill=gray!8,
                      minimum height=0.75cm, minimum width=2.4cm, align=center,
                      font=\small},
    loopnode/.style={circle, draw=black!40, fill=teal!8,
                 minimum size=1.55cm, align=center, font=\footnotesize,
                 inner sep=1pt},
    output/.style={rectangle, rounded corners=2pt, draw=black!40, fill=gray!6,
                   minimum height=0.6cm, align=center, font=\footnotesize},
    cmdlabel/.style={font=\scriptsize\itshape, text=black!55, align=left},
    arrowstyle/.style={->, thick, black!55},
]

\node[actor] (math) at (0, 0) {Researcher};
\node[component, right=1.4cm of math, fill=orange!8, draw=orange!50!black!50]
    (instructions) {Instruction File\\[-1pt]{\scriptsize\texttt{\{CLAUDE,GEMINI,AGENTS\}.md}}};
\node[actor, right=1.4cm of instructions, fill=teal!8, draw=black!40]
    (agent) {CLI Agent};
\node[component, right=1.4cm of agent, fill=yellow!12, draw=yellow!55!black]
    (sandbox) {Sandbox\\[-1pt]{\scriptsize Python, \LaTeX, Git, GPU}};

\draw[arrowstyle] (math) -- node[above, font=\scriptsize] {prompts} (instructions);
\draw[arrowstyle] (instructions) -- node[above, font=\scriptsize] {governs} (agent);
\draw[arrowstyle] (agent) -- node[above, font=\scriptsize] {runs in} (sandbox);

\draw[arrowstyle] (agent.north) to[out=120, in=60, looseness=0.6]
    node[above, font=\scriptsize] {reports back} (math.north);

\def\cx{5.5}
\def\cy{-4.2}
\def\r{2.5}

\node[loopnode] (s1) at ({\cx+\r*sin(0)},   {\cy+\r*cos(0)})   {\textbf{1}\\Explore};
\node[loopnode] (s2) at ({\cx+\r*sin(45)},  {\cy+\r*cos(45)})   {\textbf{2}\\Plan};
\node[loopnode] (s3) at ({\cx+\r*sin(90)},  {\cy+\r*cos(90)})   {\textbf{3}\\Implement};
\node[loopnode] (s4) at ({\cx+\r*sin(135)}, {\cy+\r*cos(135)})  {\textbf{4}\\Evaluate};
\node[loopnode] (s5) at ({\cx+\r*sin(180)}, {\cy+\r*cos(180)})  {\textbf{5}\\Analyze};
\node[loopnode] (s6) at ({\cx+\r*sin(225)}, {\cy+\r*cos(225)})  {\textbf{6}\\Record};
\node[loopnode] (s7) at ({\cx+\r*sin(270)}, {\cy+\r*cos(270)})  {\textbf{7}\\Commit};
\node[loopnode] (s8) at ({\cx+\r*sin(315)}, {\cy+\r*cos(315)})  {\textbf{8}\\Iterate};

\foreach \i/\j in {s1/s2, s2/s3, s3/s4, s4/s5, s5/s6, s6/s7, s7/s8, s8/s1} {
    \draw[arrowstyle] (\i) -- (\j);
}

\node[font=\small\bfseries, align=center, text=black!60] at (\cx, \cy)
    {Experiment\\[-1pt]Loop};

\node[cmdlabel, anchor=west] (c28) at ({\cx+\r*sin(45)+1.1}, {\cy+\r*cos(45)})
    {\hyperref[cmd:VI]{VI}: one variable\\\hyperref[cmd:VIII]{VIII}: bound expectations};
\draw[dotted, black!35] (s2) -- (c28);

\node[cmdlabel, anchor=west] (c5) at ({\cx+\r*sin(90)+1.1}, {\cy+\r*cos(90)})
    {\hyperref[cmd:V]{V}: make it work};
\draw[dotted, black!35] (s3) -- (c5);

\node[cmdlabel, anchor=west] (c27) at ({\cx+\r*sin(135)+1.1}, {\cy+\r*cos(135)})
    {\hyperref[cmd:II]{II}: honest evaluation\\\hyperref[cmd:VII]{VII}: three tiers};
\draw[dotted, black!35] (s4) -- (c27);

\node[cmdlabel, anchor=north west] (c10) at ({\cx+\r*sin(180)+0.9}, {\cy+\r*cos(180)-0.3})
    {\hyperref[cmd:X]{X}: verify before claiming};
\draw[dotted, black!35] (s5) -- (c10);

\node[cmdlabel, anchor=east, align=right] (c39) at ({\cx+\r*sin(225)-1.1}, {\cy+\r*cos(225)})
    {\hyperref[cmd:III]{III}: verify citations\\\hyperref[cmd:IX]{IX}: record everything};
\draw[dotted, black!35] (s6) -- (c39);

\node[cmdlabel, anchor=east, align=right] (c14) at ({\cx+\r*sin(315)-1.1}, {\cy+\r*cos(315)})
    {\hyperref[cmd:I]{I}: keep promises\\\hyperref[cmd:IV]{IV}: complete all work};
\draw[dotted, black!35] (s8) -- (c14);

\draw[arrowstyle, very thick, black!45] (agent.south) -- (s1);

\node[output] (git) at ({\cx+\r*sin(270)-2.0}, {\cy+\r*cos(270)}) {Git history};
\draw[arrowstyle, black!35] (s7) -- (git);

\pgfmathsetmacro{\reportX}{\cx+\r*sin(225)}
\node[output, minimum width=2.8cm] (report) at (\reportX, \cy-\r-1.3)
    {\texttt{report.tex}\,/\,\texttt{TODO.md}};
\draw[arrowstyle, black!35] (s6.south) -- (report.north);

\node[font=\footnotesize\itshape, text=orange!0!black, anchor=north]
    at (\cx, \cy-\r-2.0)
    {All steps governed by the Ten Commandments in \Cref{sec:commandments}.};

\end{tikzpicture}%
}}
  \caption{Overview of the agentic research framework.
  \textbf{Top:} The researcher writes a persistent instruction file that governs the CLI agent operating within a sandboxed environment.
  \textbf{Bottom:} Each experiment follows an eight-step loop.}
  \label{fig:framework}
  \end{figure}

\subsection{The Ten Commandments}
\label{sec:commandments}

At the core of our framework are the lessons we distilled through experimentation into ten commandments that apply independently of the specific domain and research problem. They form a major part of the instructions given to the agent. The full instructions are available in our repository.

In deriving the ten commandments through continuous improvement of the agent's behavior, we followed three guiding principles:
(1)~\emph{explicit over implicit:} language models follow instructions literally; implicit expectations (``obviously you should record your results'') are reliably violated, so every important behavior must be stated as a rule;
(2)~\emph{falsifiable over aspirational:} ``be rigorous'' is not a commandment, ``change exactly one variable per experiment'' is, allowing both human and agent to assess compliance;
(3)~\emph{failure-driven over theory-driven:} every commandment exists because we observed a specific failure mode in practice, not because it seemed theoretically desirable.

The commandments are grouped into categories, each addressing a specific aspect of the research process. Below, we state each rule and describe the failure mode it addresses. We present slightly shortened versions for brevity; the full prompts are available on GitHub. At the implementation level, each commandment is a prompt-engineering directive; we found that naming and structuring these behaviors as explicit rules makes them significantly easier to maintain, debug, and iterate on.
\subsubsection{Integrity and Trust}
The following three commandments address the integrity of the agent's promises and announced actions.
\phantomsection\label{cmd:I}
\begin{commandmentbox}[I. Never Break a Promise]
If you say ``I will do X,'' do it.
Under-promise, over-deliver.
\end{commandmentbox}

\noindent\textbf{Failure mode:}
In early experiments, the agent frequently stated intentions (``I will now run the full evaluation'') and then skipped steps or moved on to different tasks. After adding the commandment, the agent either follows through on all stated tasks or states upfront which tasks will be deferred and why.

\phantomsection\label{cmd:II}
\begin{commandmentbox}[II. Never Manipulate Evaluation]
Do not change metrics, test sets, fixed hyperparameters, or problem definitions.
Do not hardcode results or cherry-pick seeds.
\end{commandmentbox}

\noindent\textbf{Failure mode:}
The agent subtly changes evaluation conditions to make results look better. The LLM may adjust evaluation parameters ``helpfully'' to reach its goal, but this is not a genuine improvement. For instance, the agent changed the number of evaluation samples to ``speed up evaluation'', which happened to produce better metrics and created an unfair advantage over baseline methods.

\phantomsection\label{cmd:III}
\begin{commandmentbox}[III. Never Fabricate Citations]
Every bibliography entry must be verified against the actual source before adding it.
Search for the paper via web search.
Confirm the \emph{exact} title, \emph{full} author list, year, venue, and identifier from the source.
If you cannot find the paper, do not guess.
Never write a citation from memory alone.
\end{commandmentbox}

\noindent\textbf{Failure mode:}
This commandment aims to address a well-known limitation of these systems: they hallucinate plausible but incorrect bibliographic entries.

\subsubsection{Autonomy and Efficiency}
A major problem we encountered was that, despite having a long todo-list of potential tasks and experiments, the agent consistently stopped to ask whether it should continue. The following two commandments aim at maximizing productive work within each session.

\phantomsection\label{cmd:IV}
\begin{commandmentbox}[IV. Complete All Autonomous Work Before Reporting]
Finish every task that does not need user input.
Report once with all results.
Never skip work because you estimate it ``takes too long to implement''.
\end{commandmentbox}

\noindent\textbf{Failure mode:}
The agent frequently stops to ask whether it should continue, even when the research plan specifies many more experiments that could be executed without additional input from the researcher. A related failure mode is that the agent often discards approaches because they ``would take too long to implement'' and potentially ``only have modest impact''. Modest impact aside, agents drastically underestimate their own coding speed; in fact, the implementation typically takes less than a minute. The only valid time concern is actual compute runtime measured in days.

\phantomsection\label{cmd:V}
\begin{commandmentbox}[V. Make It Work Before Moving On]
An experiment crash is a bug, not a bad idea.
Do not discard methods because of implementation failures.
Investigate, fix, and re-run.
\end{commandmentbox}

\noindent\textbf{Failure mode:}
When encountering an implementation failure, agents often claim that the approach ``doesn't work'' and move on to an alternative. In practice, however, most of these crashes are simple bugs that can be fixed easily. For instance, when hitting an out-of-memory error, the agent concluded that the method ``doesn't scale''. Upon further investigation, it found an unnecessary materialization of a memory-intensive matrix, replaced it, and the method ran successfully, yielding significant improvements over the baseline.

\subsubsection{Scientific Rigor}
The following three commandments ensure that the agent follows the norms of scientific practice.

\phantomsection\label{cmd:VI}
\begin{commandmentbox}[VI. One Variable per Experiment]
Change exactly one thing per experiment.
If two things change and the metric improves, you cannot know which helped.
\end{commandmentbox}

\noindent\textbf{Failure mode:}
If one experiment is successful and the agent has an idea for further improvement, it is often tempted to combine both the successful change and the new idea simultaneously in the next experiment. This makes it impossible to determine which change caused the improvement.

\phantomsection\label{cmd:VII}
\begin{commandmentbox}[VII. Evaluate in Tiers]
Tier~1 (seconds): does it run without crashing?\\
Tier~2 (minutes): any signal on a small subset?\\
Tier~3: full evaluation, i.e., the real metric that goes into the report.
Use small-scale runs to catch bugs only. Never draw conclusions from small-scale results.
\end{commandmentbox}

\noindent\textbf{Failure mode:}
We want the agent to iterate quickly and distinguish between trivial and meaningful improvements. Consequently, we enforce that the agent (a)~does not run a full, potentially expensive evaluation after every minor code change, and (b)~does not discard ideas based on unsuccessful small-scale runs on toy problem instances.

\phantomsection\label{cmd:VIII}
\begin{commandmentbox}[VIII. Bound Your Expectations]
Before implementing a heuristic, identify the theoretical best case, even if it is not realizable in practice.
If you are ``correcting'' something, measure how much correction is theoretically possible.
\end{commandmentbox}

\noindent\textbf{Failure mode:}
To decide whether a method is successful, it is crucial to understand a theoretical upper bound on the possible improvement. The agent often observes a small improvement and reports it as a success, without assessing proximity to the theoretical maximum.

\subsubsection{Documentation and Reproducibility}
The following two commandments ensure that the agent documents its work reproducibly. This is one of the most important categories, as it enables restarting the research process from any given point.

\phantomsection\label{cmd:IX}
\begin{commandmentbox}[IX. Record Everything]
Every experiment gets a subsection in the report: goal, hypothesis, method, results table, analysis, next steps.
Include failures. If it is not in the report, it did not happen.
Visualize, don't just describe: create plots for distributions, comparisons, and scaling.
Maintain \texttt{TODO.md} as a living checklist for open questions, unverified claims, and deferred work.
\end{commandmentbox}

\noindent\textbf{Failure mode:}
Without the rule, the agent runs experiments, observes results, and keeps them in its context window. As soon as this context window is compacted or cleared, the information is lost. At the same time, the strict rule ``if it is not in the report, it did not happen'' ensures that the agent does not mistakenly believe it has already obtained a result that was never recorded. Apart from the report, which we save as a \LaTeX{} document, we also maintain a \texttt{TODO.md} file, which is equally critical, as it prevents the agent from forgetting about open questions, unverified claims, and deferred work.

\phantomsection\label{cmd:X}
\begin{commandmentbox}[X. Verify Before Claiming]
Assume you are wrong until verified.
Write verification scripts, not just explanations.
Actively try to falsify your own claims, test edge cases, randomize inputs, search for counterexamples.
Grade claims: \emph{verified}, \emph{partially verified}, or \emph{unverified}.
\end{commandmentbox}

\noindent\textbf{Failure mode:}
Mathematical verification remains a major challenge for LLMs. We observed significant improvements when enforcing at least numerical verification of claims. For instance, the agent derives a formula whose derivation contains an error (e.g., a missing factor of two), but the results look plausible. A verification script that checks the formula against a brute-force computation on small instances catches this immediately and prevents the agent from continuing its argument on a false premise. This active falsification, i.e., the process of deliberately trying to break your own hypothesis before confirming it, often reveals the key structural insight that makes the proof work.

\subsection{Domain-specific Commandments}
\label{sec:modules}
The ten commandments presented above are intended to be universal. In addition, we found it beneficial to provide domain-specific commandments tailored to the research style of the domain, whether primarily theoretical or empirical. Beyond these broad categories, further specialization is useful: for instance, research in a specific subfield of mathematics benefits from commandments tailored to its particular challenges.

\paragraph{Domain: Compute-Intensive Research.}
For empirical projects involving GPU experiments, deep learning, or large-scale numerical simulations, we apply the following additional commandments:
\begin{itemize}[leftmargin=*, itemsep=2pt]
    \item \phantomsection\label{cmd:C1}\textbf{One experiment per GPU; use them all (C1).} Check \texttt{nvidia-smi} before every batch of work. Assign each independent experiment to its own GPU. Never leave GPUs idle when independent tasks remain.
    \item \phantomsection\label{cmd:C2}\textbf{Context window hygiene (C2).} Prefer redirecting long-running output to log files and monitoring with \texttt{tail}. Only investigate logs in detail if something looks wrong. 
    \item \phantomsection\label{cmd:C3}\textbf{Memory management (C3).} When observing out-of-memory (OOM) errors, do not conclude that the method ``does not scale''. Instead, systematically reduce memory: clear the GPU cache between experiments (\texttt{torch.cuda.empty\_cache()}), enable gradient checkpointing, or process layers sequentially instead of in parallel. Print \texttt{torch.cuda.memory\_summary()} to identify the allocation that causes the spike. Only after these mitigations fail is it valid to report a genuine scaling limitation.
    \item \phantomsection\label{cmd:C4}\textbf{Discover nodes first; dispatch independent experiments (C4).} When a multi-node Slurm allocation is active, discover available nodes at session startup and dispatch independent experiments to remote nodes via \texttt{remote-run}. Each dispatched job runs in its own container on the target node with full GPU access. Never dispatch dependent work: only experiments that are fully independent may run on remote nodes.
\end{itemize}

\paragraph{Domain: Mathematical Research.}
For theory-heavy projects involving proofs and derivations, we apply the following additional commandments:
\begin{itemize}[leftmargin=*, itemsep=2pt]
    \item \phantomsection\label{cmd:M1}\textbf{Derivations before code (M1).} Write derivations step-by-step before implementing. Cross-reference paper equations. Before implementing a new method, search for prior work to flag potential rediscovery. 
    \item \phantomsection\label{cmd:M2}\textbf{Precise notation (M2).} Use precise index notation ($G_{jj}$, not $G_j$, for diagonal elements of a matrix). Define all notation before first use; dimensions, ranges, scalar vs.\ vector vs.\ matrix. Apply the same rigor to negative results as to positive ones. 
    \item \phantomsection\label{cmd:M3}\textbf{Counterexample-first reasoning (M3).} Before attempting a proof, actively search for counterexamples: randomize inputs, test boundary cases, enumerate small instances exhaustively. If a counterexample exists, the search finds it faster than a failed proof attempt reveals the obstruction. If no counterexample survives, the search often exposes the structural property that makes the proof work. 
\end{itemize}

\section{Case Studies}
\label{sec:case_studies}
We present case studies demonstrating the framework across different research domains and integration levels.
The first three (A--C) deal with LLM-related research questions: pretraining, pruning, and quantization. The remaining three (D--F) concern mathematical research: convex optimization, combinatorial optimization, and algebraic geometry.
Each case study follows a consistent structure: domain, problem, what the agent did, results, and lessons learned.
Throughout, we include figures, screenshots, and excerpts from the agent's reports as they were produced (indicated by a thin border); minor errors or rendering artifacts are preserved and marked with [\textit{sic}] where appropriate.

\subsection{Systematic Optimizer Exploration for LLM Pretraining}
\label{sec:case_a}

This case study demonstrates the framework's core experimental loop on a computationally intensive deep learning task: systematic, single-variable experimentation across a non-trivial optimizer design space, with multiple GPUs running independent experiments in parallel.

\paragraph{Domain and problem.}
AdamW~\citep{kingma2014adam,loshchilov2017decoupled} has long been the dominant optimizer for language model pretraining. It maintains two buffers per parameter (first and second moments), requiring additional memory $2N$ compared to vanilla Stochastic Gradient Descent (SGD), where $N$ is the number of parameters.
The Muon optimizer~\citep{jordan2024muon} takes a fundamentally different approach: instead of adaptive step sizes, it computes a momentum vector $M_t = \mu M_{t-1} + G_t$ and then applies Newton-Schulz (NS) orthogonalization to approximate $UV^\top$ from the Singular Value Decomposition (SVD) of the momentum buffer $M_t=U\Sigma V^\top$, so that $W_{t+1} = W_t - \eta \cdot \mathrm{NS}(M_t)$.
This operation equalizes all singular values of the update and achieves strong results on LLM pretraining while using only $N$ additional memory units (one momentum buffer) compared to SGD, half of AdamW's $2N$.
A natural question arises: \emph{can the spare $N$ memory budget be exploited to make Muon better?}
The agent was given this open-ended research question, the codebase of \citet{semenov2025benchmarking} as a standardized LLM pretraining benchmark (124M-parameter Llama on FineWeb, 10{,}000 iterations), and a multi-GPU compute allocation.

\paragraph{What the agent did.}
After establishing baselines (Muon, AdamW), the agent explored modifications to the Muon update rule, changing exactly one variable per experiment (\cmdref{VI}).
The central insight was that Muon converges faster when the vector it orthogonalizes is well-conditioned: normalizing the momentum buffer before orthogonalization means the same number of iterations yields a better update.
The agent tested multiple normalization strategies, swept hyperparameters one at a time, and discovered two independent improvements: (1)~a normalization technique applied before orthogonalization, and (2)~the addition of weight decay to Muon's matrix parameters.
Weight decay is a standard regularization technique and its benefit is not surprising in itself; however, the reference codebase implemented Muon without it, and because the agent tested each modification in isolation (\cmdref{VI}), it was able to quantify this contribution separately and still identify the normalization improvement on top of it.
A zero-overhead variant requiring no extra buffer was found to achieve nearly identical results.
Following \cmdref{IX}, each of the more than 40 experiments was documented in the agent's \texttt{report.tex} with goal, hypothesis, method, results table, and analysis.

The agent also identified several independent papers exploring normalization in the context of Muon: NorMuon~\citep{li2025normuon}, AdaMuon~\citep{si2025adamuon}, and Muon+~\citep{zhang2026muon+}, each proposing a different normalization strategy.
It implemented two of these methods in its codebase and ran a detailed comparison, analyzing the theoretical and empirical differences between the approaches (\cmdref{V}).
The existence of multiple concurrent works exploring the same design space underscores the need to carefully characterize how the agent's approach relates to and differs from each of them.
While the agent conducted thorough literature searches, we cannot guarantee that its specific combination of modifications is truly novel.
Accordingly, we keep the presentation at a high level and view these results primarily as initial directions to build on: the experiments are limited to a single architecture and dataset, and a full comparison across model scales, training setups, and concurrent methods would be necessary to draw any definitive conclusions.
A standalone publication would further require a more in-depth prior-art investigation to establish precisely which aspects, if any, are new.

\begin{figure}[t]
\centering
{\fboxsep=4pt\fboxrule=0.4pt\fcolorbox{black!50}{white}{\includegraphics[width=0.83\textwidth]{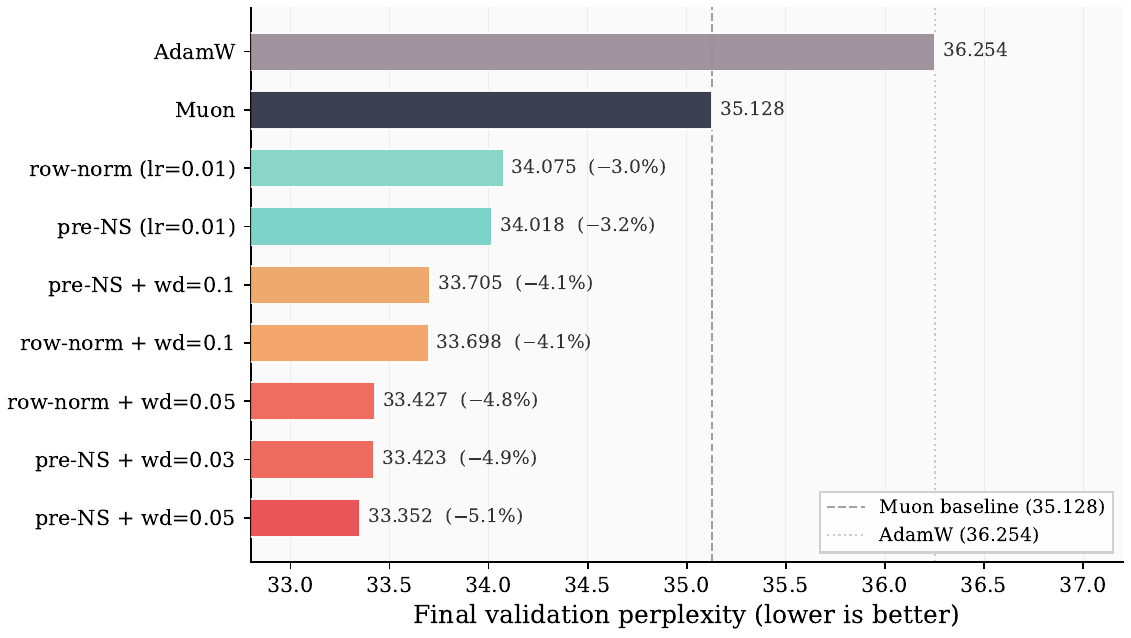}}}
\caption{Final validation perplexity [\textit{sic}] from the agent's report in \Cref{sec:case_a}. Lower is better. The dashed line marks the Muon baseline; the agent's modifications achieve ${\sim}5\%$ improvement over Muon and ${\sim}8\%$ over AdamW.}
\label{fig:case_a_ppl}
\end{figure}

\begin{figure}[t]
\centering
{\fboxsep=4pt\fboxrule=0.4pt\fcolorbox{black!50}{white}{\includegraphics[width=0.98\textwidth]{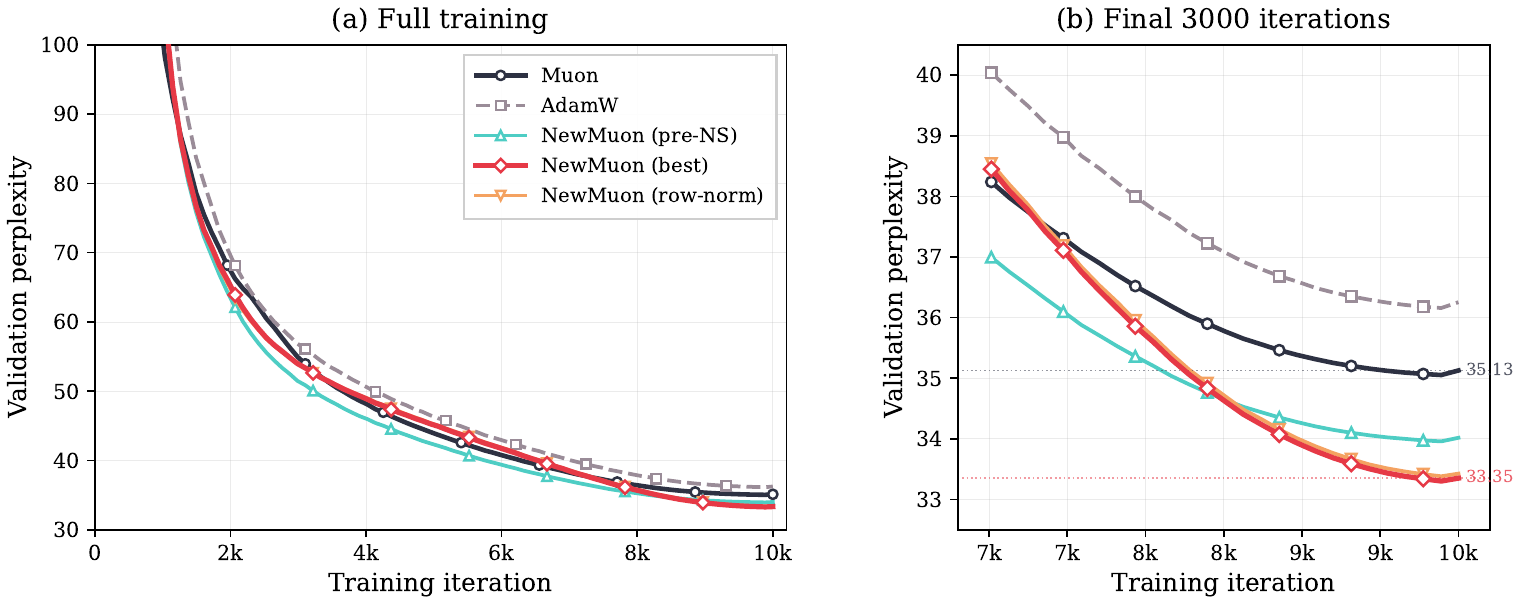}}}
\caption{Training curves [\textit{sic}] from the agent's report in \Cref{sec:case_a}. Left: full training run. Right: final 3{,}000 iterations (zoomed). The agent's optimizer modifications consistently outperform both Muon and AdamW baselines throughout training, not only in the final iterations. Note that here, the agent named the new method NewMuon, which is inconsistent with the naming in \Cref{fig:case_a_ppl}.}
\label{fig:case_a_curves}
\end{figure}

\paragraph{Results.}
Across more than 40 experiments documented in the agent's \texttt{report.tex}, the best configuration achieved a ${\sim}5\%$ improvement in validation perplexity over Muon (and ${\sim}8\%$ over AdamW) at the same $2N$ memory budget as AdamW (\Cref{fig:case_a_ppl}).
The two improvements are nearly additive: normalization alone provides ${\sim}3\%$, weight decay alone ${\sim}2\%$, and the combination ${\sim}5\%$ (\Cref{fig:case_a_curves}).
The zero-overhead variant achieves ${\sim}4.8\%$ improvement at the \emph{same} $N$ memory footprint as baseline Muon, within a fraction of a perplexity point of the full method. Results were replicated across random seeds and a broader hyperparameter sweep.

\paragraph{Lessons learned.}
The one-variable-at-a-time commandment (\cmdref{VI}) was critical in this design space: the agent discovered that normalization and weight decay provide independent, nearly additive improvements only because it tested each in isolation before combining them.
A $2 \times 2$ factorial ablation (normalization $\times$ weight decay) confirmed the near-additivity, which would have been obscured by testing them jointly from the start.
An interesting aspect of the agent's research behavior is that, while the task explicitly granted an extra $N$ memory budget, the agent proactively explored whether the same gains could be achieved without it, and found a zero-overhead variant that nearly matched the full method at the same $N$ memory footprint as baseline Muon.
The entire session ran for over twenty hours without human intervention.
With multiple GPUs available, the agent ran independent experiments in parallel (one per GPU, \cmdref{C1}); the framework's multi-node dispatch capability (\Cref{sec:modules}) enables large-scale concurrent experiments across compute nodes.
Despite the long wall-clock time, actual token consumption remained modest: most time was spent waiting for training runs to finish while the agent redirected output to log files and monitored progress with lightweight commands (\Cref{fig:terminal_session}), as encouraged by \cmdref{C2}.
The framework's emphasis on literature verification (\cmdref{III}) prompted the agent to proactively search for related work, identify concurrent papers, and implement their methods for comparison. While this is a useful first step, the limitations noted above show that such automated searches are not a substitute for the thorough prior-art investigation a human researcher would conduct before claiming novelty or asserting that the resulting method truly outperforms concurrent approaches.

\subsection{Weight Reconstruction in Large Language Model Pruning}
\label{sec:case_b}

This case study illustrates a characteristic side effect of the agentic framework we propose: the agent was tasked with one research objective and \emph{discovered} a different, more effective technique along the way (i.e., we observed \emph{serendipity}).

\paragraph{Domain and problem.}
Pruning large language models (LLMs) reduces memory and compute costs by zeroing out weights, i.e., selecting a binary sparsity mask $M \in \{0,1\}^{d_{\text{out}} \times d_{\text{in}}}$ per weight matrix \citep[cf., e.g.,][]{zimmer2023perp,frantarSparsegptMassiveLanguage2023,sunSimpleEffectivePruning2024}.
The constraints on $M$ determine the sparsity pattern and, with it, the potential for hardware acceleration: unstructured sparsity removes arbitrary individual weights~\citep{han2015learning, Zimmer2023, zimmer2023sparse,ZimmerSpiegelPokutta+2025+137+168}, while semi-structured patterns such as $N{:}M$~\citep{mishra2021accelerating, zhangPlugandPlayEfficientPosttraining2023, lasby2025compressed} impose structure that is more amenable to hardware acceleration.
The core challenge across all settings is \emph{mask selection}: choosing which weights to zero out so that the pruned network's output remains close to the original~\citep{roux2025dontbegreedyjustrelax,zimmerSparseSwapsTractableLLM2026}.
Once a mask is fixed, the pruned model's performance degrades compared to the dense original; one way to counteract this is \emph{weight reconstruction}, i.e., adjusting the surviving weights to compensate for the removed connections~\citep{frantarSparsegptMassiveLanguage2023}.
Calibration data is drawn from C4~\citep{raffel2020exploring}; quality is measured by perplexity on the WikiText~\citep{merity2016pointer} test set (lower is better).

The project started with a concrete task: we had developed a pruning approach that aimed to find better masks, but it produced inconsistent results, sometimes failing catastrophically.
The agent was provided with an existing codebase containing implementations of several pruning methods and the \LaTeX{} derivation of our approach, and instructed to analyze why it failed, fix or replace the method, and empirically beat a set of baselines~\citep{sunSimpleEffectivePruning2024, zhangPlugandPlayEfficientPosttraining2023} at 60\% sparsity.

\paragraph{What the agent did.}
The agent first established that the existing approach was mathematically flawed and could not be repaired.
While analyzing \emph{why} it failed, the agent studied how pruning distorts the post-layer activations of each weight matrix and observed a severe imbalance: some rows lose over 50\% of their activation-weighted output magnitude while others lose less than 10\%.
This byproduct of debugging led the agent to propose a simple post-pruning weight correction that restores the activation balance across rows and columns.
Following \cmdref{VIII}, the agent first computed an oracle bound via least-squares reconstruction to determine the theoretical limit, then validated the new method through the tiered evaluation protocol (\cmdref{VII}) across five model scales.

\begin{figure}[t]
\centering
{\fboxsep=4pt\fboxrule=0.4pt\fcolorbox{black!50}{white}{%
\begin{minipage}[t]{0.47\textwidth}\centering
\vspace{0pt}
\includegraphics[width=\textwidth]{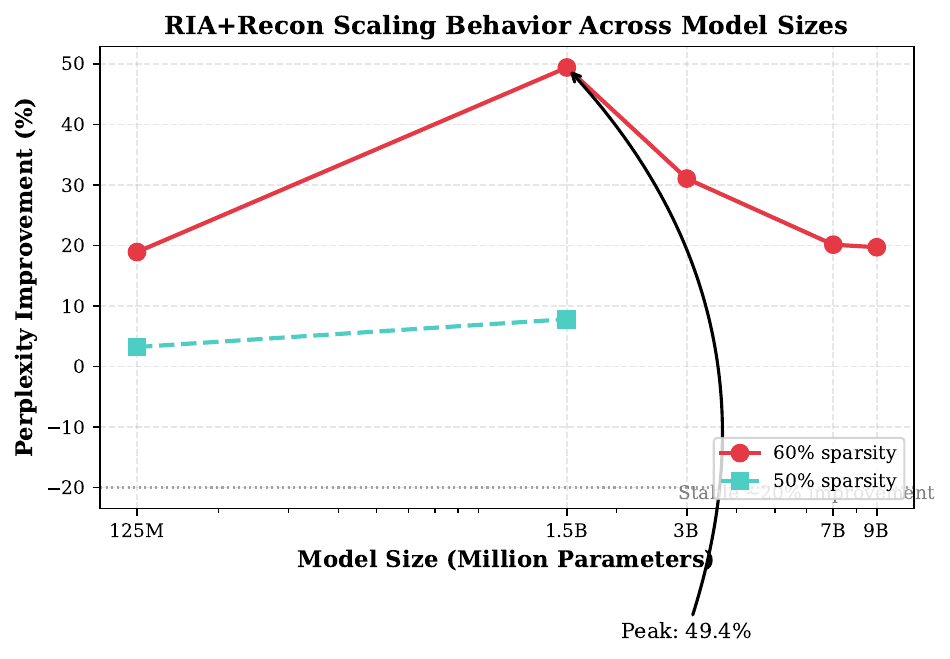}
\end{minipage}
\hfill
\begin{minipage}[t]{0.47\textwidth}\centering
\vspace{0pt}
\includegraphics[width=\textwidth]{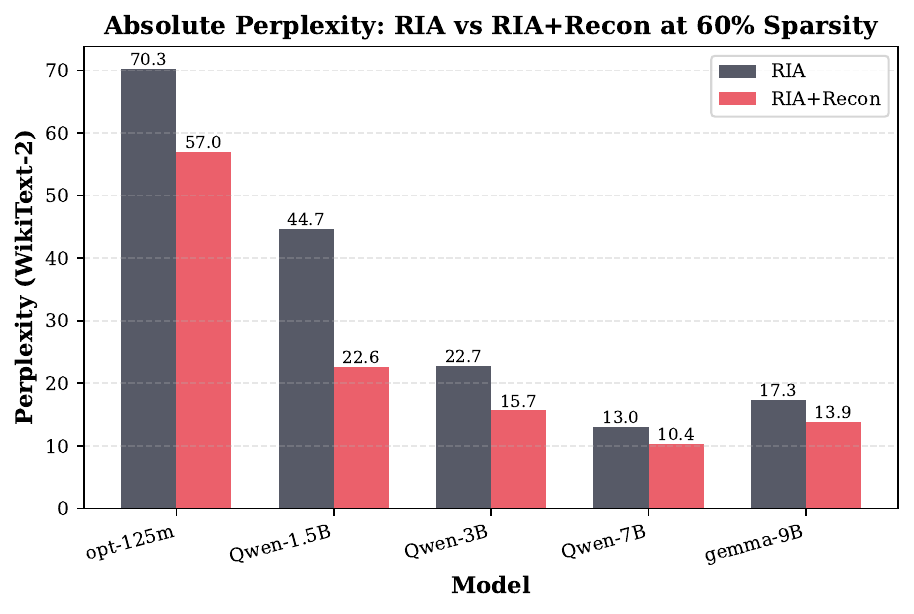}
\end{minipage}}}
\caption{Plots [\textit{sic}] from the agent's report for \Cref{sec:case_b}, produced by the agent. Left: relative perplexity improvement vs.\ model size. Right: absolute perplexity comparison showing that the weight reconstruction method consistently outperforms the baseline across all tested model sizes.}
\label{fig:case_b_scaling}
\end{figure}

\paragraph{Results.}
The method consistently reduces perplexity by 18--50\% across five model scales (125M to 9B parameters), three architectures (OPT, Qwen, Gemma), and two pruning methods (RIA, Wanda).
It requires only 10 lines of code, adds less than 1\% computational overhead, and needs no hyperparameter tuning.
The oracle comparison shows that this simple heuristic captures 92\% of the improvement achievable by full least-squares reconstruction, leaving little room for more sophisticated approaches.
Across 27 experiments documented in the agent's report, the improvements are robust and transfer to every model and pruning method tested.
\Cref{fig:case_b_scaling} shows the scaling behavior across model sizes, reproduced [\textit{sic}] from the agent's report; note, for instance, that the 50\% sparsity line in the left panel ends at 1.5B because the agent found the 60\% setting more promising and did not complete the remaining experiments.

\paragraph{Lessons learned.}
The original task was to fix a broken pruning mask; the actual outcome was a novel weight reconstruction method.
The commandments forced the agent to analyze \emph{why} the approach failed rather than simply trying the next idea, and this systematic analysis led to the discovery.
Computing the oracle baseline (\cmdref{VIII}) early on established that 92\% of the theoretical optimum was already achieved, preventing wasted effort on a nearly closed gap.
Finally, several extensions showed no benefit on small models but 7--11\% improvement at 1.5--7B scale; the tiered evaluation protocol (\cmdref{VII}) caught this systematically.

\subsection{Column Ordering in LLM Quantization}
\label{sec:case_c}

This case study shows the framework operating as a systematic empirical researcher: given a well-defined design space, the agent mapped it comprehensively and discovered that the most important finding was not which method wins, but \emph{when and why} it matters.

\paragraph{Domain and problem.}
Post-training quantization compresses a pretrained language model by representing its weights in lower precision, substantially reducing the memory footprint and enabling deployment on consumer-grade hardware.
GPTQ~\citep{frantarGPTQAccuratePostTraining2023}, a widely used method, processes each weight matrix $W \in \mathbb{R}^{d_{\text{out}} \times d_{\text{in}}}$ column by column to minimize the layer-wise reconstruction error $\|(W - \hat{W})X\|_F^2$, where $\hat{W}$ denotes the quantized matrix and $X \in \mathbb{R}^{d_{\text{in}} \times n}$ are calibration activations.
Each column's rounding error is propagated to subsequent columns via the inverse of the Hessian $H = 2XX^\top \in \mathbb{R}^{d_{\text{in}} \times d_{\text{in}}}$.
The order in which columns are processed affects the final quality.
A post-publication variant known as ``act-order''\footnote{Commit \texttt{a4c3c89}, March 2023, in \url{https://github.com/IST-DASLab/gptq}.} sorts columns by descending Hessian diagonal, with the intuition that high-sensitivity columns benefit from having more subsequent columns available for error compensation.
The agent was tasked with investigating whether better orderings exist, how the effect depends on model architecture, and validating findings across model families.
Calibration data is drawn from C4~\citep{raffel2020exploring}; quality is measured by perplexity on the WikiText~\citep{merity2016pointer} test set (lower is better).

\begin{figure}[t]
\centering
\includegraphics[width=0.9\textwidth]{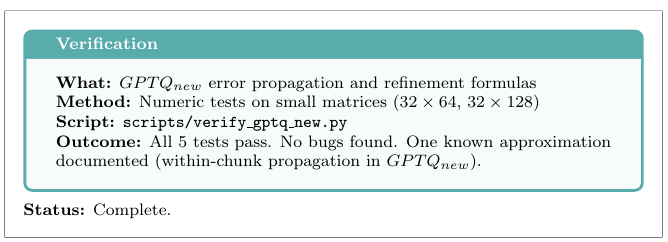}
\caption{A screenshot [\textit{sic}] from the agent's report in \Cref{sec:case_c}. Before running any benchmarks, the agent audited all error propagation and refinement formulas through numeric tests on small matrices (\cmdref{X}).}
\label{fig:case_c}
\end{figure}

\paragraph{What the agent did.}
The agent began with a mathematical analysis of why column ordering matters, then implemented and compared seven ordering strategies, first on single weight matrices, then at full model scale.
Following \cmdref{X}, it created verification scripts for all error propagation and refinement formulas before running any benchmarks (\Cref{fig:case_c}).
Cross-architecture validation (\cmdref{VII}) across five model families (Qwen, Llama, Gemma, Mistral, Yi) revealed the central finding: the ordering effect varies by more than two orders of magnitude across architectures.

\paragraph{Results.}
Column ordering is the single most impactful improvement to GPTQ, but its magnitude is entirely architecture-dependent: it reduces perplexity by 74\% on Llama-3.1-8B but only 0.1\% on Gemma-2-9B at 4-bit.
This finding would have been missed without systematic multi-architecture validation: on Qwen-1.5B alone, the effect is 20\%, giving no indication that it ranges from 0.1\% to 74\% across architectures.
Among the seven ordering strategies tested, alternatives that incorporate the quantization error magnitude alongside column sensitivity occasionally outperformed act-order (e.g., at 3-bit on certain architectures), but no single strategy dominated consistently across all architectures and bit widths.
Nine of the 24 experiments produced negative results, each documented with the same rigor as positive ones (\cmdref{IX}): many approaches failed because GPTQ's error propagation via Ordinary Least Squares (OLS) already minimizes the correlations these methods would exploit.
A critical implementation bug in group quantization was caught because the agent investigated a failure rather than abandoning the method (\cmdref{V}): pre-computing scale parameters from initial instead of error-propagated weights produced catastrophic results (perplexity 437 vs.\ 9.22 after the fix).
The agent's report documents all 24 experiments and 11 key findings.

\paragraph{Lessons learned.}
The negative results (9 of 24 experiments) were more informative than the positive ones: each failure clarified \emph{why} simpler methods work, revealing that GPTQ's OLS-based error propagation already handles what sophisticated alternatives attempt.
With four GPUs, the agent ran independent model evaluations in parallel (one per GPU, \cmdref{C1}), efficiently covering five model families with multiple configurations each.
The ``Make It Work'' commandment (\cmdref{V}) prevented a false negative: group quantization initially appeared broken on Llama, but investigation revealed a subtle implementation bug whose fix turned a catastrophic failure into the best result.

\subsection{Tight Lower Bounds for Frank-Wolfe on Uniformly Convex Sets}
\label{sec:case_d}

This case study demonstrates the framework on a problem in convex optimization, where the agent's primary output is the proof of a new theorem.
Unlike the computational and empirical case studies, the research here required sustained interaction between numerical exploration and theoretical development: the agent discovered the correct proof strategy through systematic experimentation before formalizing it.

\paragraph{Domain and problem.}
The Frank-Wolfe (FW) algorithm minimizes a smooth convex function over a convex constraint set using only a linear minimization oracle (LMO).
On strongly convex sets, the known $\mathcal{O}(1/T^2)$ upper bound was recently shown to be tight: \citet{halbey2026lower} gave a lower bound for vanilla FW in dimension~2 by analyzing the dynamics of the iterates on a worst-case instance. Shortly after, \citet{grimmer2026lower} proved an information-theoretic lower bound in the high-dimensional setting for a broad class of LMO-based algorithms.

For uniformly convex sets of order $p > 2$ (e.g., $\ell_p$-balls), \citet{kerdreux2021projection} established an upper bound of $\mathcal{O}(1/T^{p/(p-1)})$, but no matching lower bound was known.
The goal was to prove lower bounds for the uniformly convex setting based on the techniques used by \citet{halbey2026lower} or \citet{grimmer2026lower}.

\paragraph{What the agent did.}
The agent began by studying both existing lower-bound techniques and attempting to generalize the high-dimensional construction by \citet{grimmer2026lower} to $\ell_p$-balls.
This did not succeed: the construction relies on decomposing strongly convex sets as intersections of shifted Euclidean balls, and the agent did not find a direct analogue for uniformly convex sets of order $p > 2$.                                                                                                          
Following \cmdref{IX}, the agent documented this negative result and pivoted to the alternative approach of \citet{halbey2026lower}, which analyzes the FW iterates directly on a worst-case instance.
The agent derived the FW dynamics on $\ell_p$-balls in closed form and verified each component numerically (\cmdref{X}).
Experiments across multiple values of $p$ revealed that the iterates alternate in sign and settle onto a low-dimensional curve whose shape can be characterized analytically, which suggested the right proof strategy.
The agent first estimated the key constants numerically, then derived them in closed form, and finally assembled a rigorous proof for $p \ge 3$ with explicit convergence rates.
Each proof step was accompanied by Julia verification scripts using \texttt{BigFloat} arithmetic, totaling over 30 individual checks.
The case $p \in (2,3)$ was identified as qualitatively different: sign alternation breaks down intermittently, and the proof technique does not apply.

\paragraph{Results.}
The main result establishes a lower bound of $\Omega(1/T^{p/(p-1)})$ for vanilla FW on $p$-uniformly convex sets for any $p \ge 3$, matching the upper bound of \citet{kerdreux2021projection} and resolving the open question for this regime.
The proof provides explicit convergence constants, all verified numerically to $< 0.2\%$ relative error.
The case $p \in (2,3)$ remains open: numerical evidence supports the same rate, but the proof technique does not extend.

\begin{figure}[ht]
\centering

{\fboxsep=4pt\fboxrule=0.4pt\fcolorbox{black!50}{white}{\includegraphics[width=0.98\textwidth]{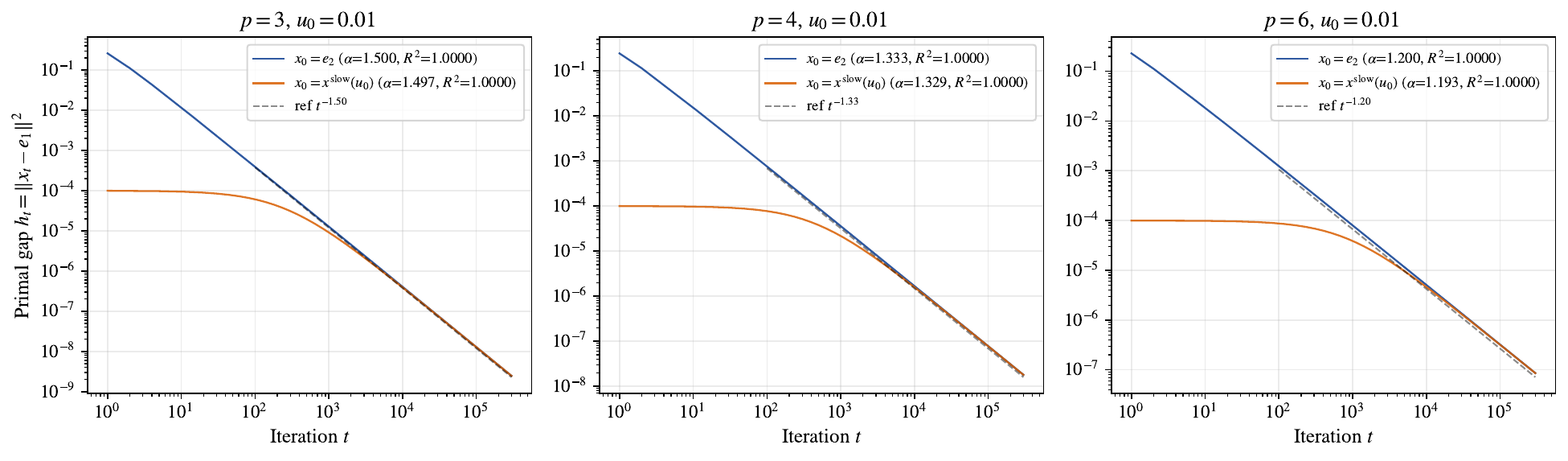}}}
\caption{A plot [\textit{sic}] from the agent's report: Log-log convergence of $\|x_t-e_1\|^2$ for $p\in\{3,4,6\}$ starting from $x_0=e_2$ (blue) and from $x_0=x_0^{\mathrm{slow}}(10^{-2})$ (orange) where $x_0^{\mathrm{slow}}$ is the worst-case initialization from the proof and $\alpha$ is the fitted coefficient of $t^{-\alpha}$.}
\label{fig:explicit-slow-init}
\end{figure}

\paragraph{Lessons learned.}
The correct proof strategy emerged from the agent's numerical exploration: patterns observed in the iterates suggested the right analytical approach, and the key constants were first estimated computationally before being derived in closed form.
This ``conjecture from computation, then prove'' loop, enabled by the framework's emphasis on creating verification scripts alongside every mathematical claim (\cmdref{X}), is a natural workflow for this type of problem.
The failed generalization of \citet{grimmer2026lower} was equally informative: it helped us understand which parts of the proof are hard to extend to the uniformly convex setting, guiding the pivot to the successful approach. Following \cmdref{IX}, this negative result was documented thoroughly.

\subsection{Multi-Variable Dual Tightening for Mixed-Integer Optimization}
\label{sec:case_e}

This case study demonstrates the framework in combinatorial optimization. Its main contribution is a multi-variable generalization of dual tightening, together with a prototype implementation in the Boscia solver. The case study spans the full research cycle: deriving the result, proving it, implementing it, and evaluating it computationally.

\paragraph{Domain and problem.}
Boscia~\citep{hendrychConvexMixedintegerOptimization2025} is a Frank-Wolfe-based branch-and-bound solver for mixed-integer nonlinear optimization over polytopes ($\min_{x \in X \cap \mathbb{Z}^J} f(x)$ with $f$ smooth convex), where $X \subseteq \R^n$.
A key pruning mechanism is \emph{dual tightening}.
At a relaxed solution $x^t$ with gradient $g = \nabla f(x^t)$ and Frank-Wolfe dual gap $\gamma(x^t) = \max_{v \in X} \langle g, x^t - v\rangle$, convexity implies that any feasible point $x \in X$ with objective value at most some upper bound $\mathrm{UB}$ (e.g., from an incumbent) satisfies $g_j(x_j - \ell_j) \le \mathrm{RHS}$ for each variable~$j$, where $\mathrm{RHS} \coloneqq \mathrm{UB} - f(x^t) + \gamma(x^t)$ and $\ell_j$ is the lower bound of~$x_j$.
This allows variables to be fixed one at a time.
The project investigated whether this extends to \emph{subsets}: for a set $S$ of variables at their lower bounds, $\sum_{j \in S} g_j(x_j - \ell_j) \le \mathrm{RHS}$, so when the combined gradient contribution exceeds the budget, a \emph{conflict constraint} prevents all variables from simultaneously deviating from their current bounds.
For binary variables, a pairwise conflict $g_i + g_j > \mathrm{RHS}$ implies $x_i + x_j \le 1$ (a conflict graph edge); higher-order conflicts (triples, quadruples) capture interactions that pairwise constraints miss.
The goal was to derive the mathematical result, implement it as a conflict graph with constraint propagation integrated into Boscia via callbacks, and benchmark on a diverse set of Mixed-Integer Nonlinear Programming (MINLP) instances.

\begin{figure}[t]
\centering
\includegraphics[width=0.8\textwidth]{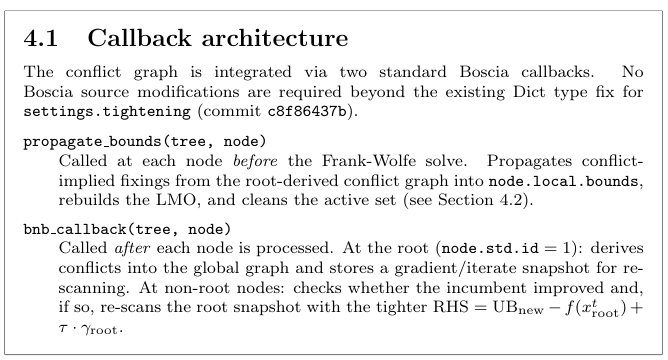}
\caption{A screenshot [\textit{sic}] from the agent's report: The callback architecture in \Cref{sec:case_e}. The conflict graph is integrated into Boscia via two standard callbacks, \texttt{propagate\_bounds} (before each Frank-Wolfe solve) and \texttt{bnb\_callback} (after each node), without modifying Boscia's source code.}
\label{fig:boscia_callback}
\end{figure}

\paragraph{What the agent did.}
The agent started from Boscia's existing single-variable dual tightening result (Theorem~3 of~\citet{hendrychConvexMixedintegerOptimization2025}), identified the natural generalization via the convexity inequality, and formulated and proved a multi-variable dual tightening theorem with corollaries for pairwise and higher-order binary conflicts.
Before implementation, the agent first tried to verify the proof both symbolically, using \texttt{Symbolics.jl} with 2{,}387 checks, and numerically, using a script that exhaustively enumerated all $2^n$ feasible points for small instances (487 checks). 
This verification caught an error in the initial derivation: the bound for the at-least set constraint had been inverted, which would have led to overly aggressive fixings for upper-bound variables.
The agent then implemented a \texttt{ConflictGraph} data structure with constraint propagation and integrated it into Boscia via two \emph{callbacks} (\Cref{fig:boscia_callback}), requiring no source modifications beyond fixing a pre-existing \texttt{Dict} type bug.
A key design decision made by the agent was to derive conflicts only at the root node. Because these conflicts use the global Frank-Wolfe gap, they remain valid throughout the search tree, but are more conservative than conflicts derived locally at child nodes. 
The agent also explored tighter child-node conflicts, but early tests suggested that the additional overhead and numerical instability were not worth the potential gain.

\paragraph{Results.}
Across 33 instances in six problem categories ($n = 12$ to $n = 300$, 10-minute time limit), partition-constrained instances show the strongest improvement (up to 18.9\% node reduction, from $127$ to $103$ nodes on a 48-variable instance), where partition constraints create tight cross-block coupling that the conflict graph captures.
The root-only design is deliberately conservative, and most instances show 0\% node reduction because the root budget is loose. However, this guarantees correctness, which is critical for an exact mixed-integer convex optimization solver, and all 33 instances produce identical optimal objectives in both modes.
As expected, separable quadratic instances show no benefit because diagonal objectives create no cross-variable coupling, confirming the theoretical prediction.

\paragraph{Lessons learned.}
This case study shows that the framework is effective for projects that combine theorem proving, verification, implementation, and experiments in a single workflow.
The verification-first approach (\cmdref{X}) was crucial for the overall correctness. It caught the inverted at-least bound bug before it entered the experiment phase.
The negative results were useful as well. The lack of improvement on separable instances matched the theory, while the $26\times$ overhead on a sparse regression instance with 150 indicator variables exposed a concrete bottleneck and pointed to straightforward fixes, including better data structures and a cap on propagated conflicts. Following \cmdref{IX}, these outcomes were all documented in the report, which made the evaluation more transparent and more useful for guiding future improvements.

\subsection{Finding Maximal Real Solutions in $K_7$ Power Networks}
\label{sec:case_f}

This case study shows the framework operating as a computational scientist for discovery. Starting from a published method for characterizing typical behavior, the agent reconstructed the pipeline and repurposed it for directed extremal search, discovering an improved lower bound.

\paragraph{Domain and problem.}
Electrical power grids can be modeled as networks of buses connected by transmission lines, where the physics imposes a system of polynomial equations whose real solutions correspond to feasible operating states. Solutions to these power flow equations define the operating points of the network and underpin decisions ranging from long-term planning and capital investment to day-to-day resource scheduling, market operations, and real-time stability analysis. The equations depend on tunable parameters (susceptances), which appear as coefficients in the system.
This motivates a natural structural question, raised explicitly by \citet{lindberg2020}: \emph{for a fixed network topology, what is the maximum number of feasible operating states over all parameter choices?} \citet{lindberg2020} characterized the \emph{distribution} of solution counts for several topologies, including $K_7$ (seven buses, every pair connected), using a continuation pipeline orders of magnitude faster than naive solving. However, they did not target extremal instances, i.e., those with a maximal number of real solutions, explicitly. 
Our goal is therefore to adapt the sampling technique from \citet{lindberg2020} to identify parameter settings that yield extremal instances.

\begin{figure}
  \centering
  {\fboxsep=4pt\fboxrule=0.4pt\fcolorbox{black!50}{white}{\includegraphics[width=0.58\textwidth]{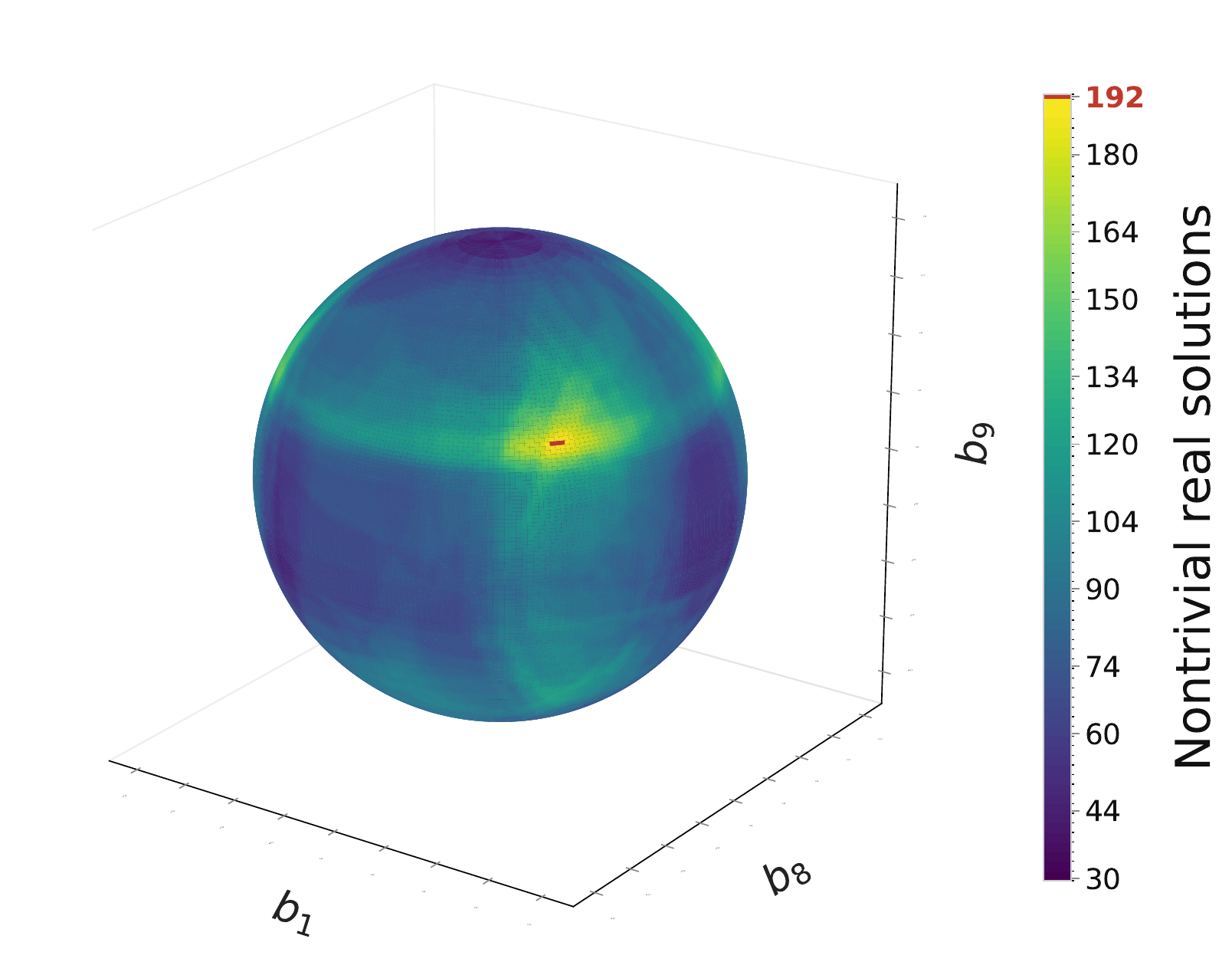}}}
  \caption{A plot [\textit{sic}] from the agent's report: a three-parameter slice of the 21-dimensional \(K_7\) susceptance space, obtained by varying \(b_1\), \(b_8\), and \(b_9\) while fixing the remaining 18 parameters at the values of the best-found instance. Each point is colored by the number of nontrivial feasible operating states. Although the color map appears nearly continuous, it represents discrete solution counts and reveals a localized high-count region around the 192-solution configuration. This suggests that the best-found parameter vector lies in a small but open region of parameter space rather than at an isolated point.}
  \label{fig:k7_sphere_slice}
\end{figure}

\paragraph{What the agent did.}
The agent first reconstructed the pipeline of \citeauthor{lindberg2020}, which was a nontrivial task. Reproducing the published results required several rounds of refinement to align the implementation with the paper's symmetry conventions, parameterization choices, and solution-counting bookkeeping. Once this baseline was validated, the agent adapted the pipeline from sampling to extremal search.
To explore the parameter space effectively, the agent combined several heuristic search strategies, including hill climbing, simulated annealing, and warm starts from the best susceptance vectors found so far. These methods were used iteratively to bias the search toward regions of parameter space with unusually large numbers of real solutions, with each successful run informing the next.

\paragraph{Results.}
Random sampling of 1.4 million parameter vectors, following the original paper's sampling protocol, found at most 120 (nontrivial) feasible states. Targeted search instead identified a parameter vector with 192 feasible states. 
The agent also perturbed this parameter vector to verify that the 192-solution count is not confined to an isolated parameter point, but persists in a neighborhood of parameter space.
\Cref{fig:k7_sphere_slice} supports this interpretation by showing that, when only \(b_1\), \(b_8\), and \(b_9\) are varied and the remaining 18 parameters are fixed, the 192-solution configuration lies in a small region with constant solution count.
The maximum real solutions problem for $K_7$ remains open. However, adapting \citeauthor{lindberg2020}'s continuation pipeline for extremal search yields a substantially stronger computational lower bound.

\paragraph{Lessons learned.}
This case study highlights the importance of verifiable intermediate artifacts: published tables and solution-count distributions were essential for checking that the reconstructed pipeline matched prior work before launching the extremal search (\cmdref{X}). It also underscored the value of staged evaluation (\cmdref{VII}): because individual searches can run for hours, the agent benefited from first validating correctness on cheaper checks and only then scaling up to long-running optimization runs. 
More broadly, the study shows that the agent need not rely on an existing codebase to begin exploration.

\section{Discussion and Conclusion}
\label{sec:discussion}

We have presented a practical framework for AI-assisted research in mathematics and machine learning, organized around a taxonomy of five integration levels, an open-source framework for working with general-purpose CLI coding agents, and case studies demonstrating this framework in practice. A central claim of this paper is that effective agentic research does not require a specialized system built from scratch. Instead, it can be built around existing general-purpose agents, provided they are embedded in a disciplined and inspectable workflow.

In our setup, the agent operates with persistent instructions, a sandboxed environment, written progress reports, \texttt{TODO.md} files, and a small set of methodological rules: change one variable at a time, evaluate in stages, and verify results before reporting them, among others. In practice, these additions were sufficient to extend the agent from a tool for isolated coding tasks into a useful research collaborator for exploratory and implementation-heavy work.

Our experience suggests a simple conclusion: model capability matters, but workflow design matters just as much. These systems are only useful when their outputs can be checked and their intermediate steps revisited. This keeps the researcher responsible for direction, judgment, and verification, even when substantial exploratory or technical work is delegated.

At the same time, this approach does not eliminate the need for expert oversight or final verification. In our framework, however, oversight is not reserved only for the end of the process; it is built into the workflow itself. A central requirement is that the agent must be able to test, challenge, and potentially refute its own claims through staged evaluation, intermediate checks, and explicit internal validation procedures. In our experience, these internal verification mechanisms are crucial. Without them, experiments can easily become structured to simply confirm an initial hypothesis. Final expert verification remains necessary, but it is far more reliable when supported by a workflow that already produces inspectable and continuously tested intermediate results. We emphasize that the case studies and reports do not constitute finished papers that are ready for publication, but rather records of meaningful research progress.

To make this approach usable by others, we release the instruction set, templates, and container definitions, with the broader goal of making AI-assisted research more systematic, reproducible, and accessible.

\subsection{Limitations}
\label{sec:limitations}

\paragraph{Verification.}
A fundamental limitation of our framework, shared with other agentic systems, is result verification.
Natural-language proofs remain difficult to verify and require manual inspection. 
While code is usually easier to check, subtle implementation errors can still invalidate conclusions.
Citations must also be verified carefully, since agents may hallucinate references or bibliographic details.
This is not only a technical limitation but also a matter of responsible use: researchers must invest substantial effort in verifying agent outputs, especially because such outputs may be even harder for others to assess independently.
As \citet{su2022} argue, researchers are often the best reviewers of their own papers; likewise, we argue that they are ultimately responsible for verifying the work produced by their agents.

\paragraph{Context.}
Long experimental sessions with many runs and large outputs can exceed a model's context window and trigger compaction.
Because compaction is inherently lossy, the agent may forget details from earlier in the session, revisit failed approaches, or miss important observations.
Practical mitigations include routing long outputs to log files and monitoring them with \texttt{tail}, manually invoking compaction commands such as \texttt{/compact}, and relying on persistent artifacts such as \texttt{report.tex} and \texttt{TODO.md} as re-entry points and external memory.
We also tested autonomous compaction, but found it to have no positive impact. Robust context management remains an open challenge.

\paragraph{Cost.}
Long autonomous sessions with frontier models can incur nontrivial API costs.
In practice, however, these costs are often relatively small since much of the wall-clock time in Level~4 sessions is spent waiting for experiments to finish rather than generating tokens.
Still, cost remains a meaningful limitation, particularly for long-running studies and large-scale evaluations.

\subsection{Future Directions}
\label{sec:future}

\paragraph{Extension to other domains.}
While our paper focuses on the application of our framework to machine learning and mathematical research, in principle it could be applied more broadly to other disciplines, such as physics, chemistry, economics, or the social sciences.
Adapting the framework to these settings would require domain-specific tools, evaluation protocols, and safety checks, but the general paradigm of iterative experimentation, artifact management, and human verification may transfer well beyond our current case studies.

\paragraph{More robust memory.}
Another important direction is improving how the system stores, retrieves, and updates information over long research sessions.
Better memory mechanisms could help agents maintain continuity across experiments, avoid revisiting failed approaches, and make more effective use of prior observations.
This would be especially valuable for complex projects that unfold over many iterations and generate substantial intermediate state.

\paragraph{Multi-user collaboration.}
Our setup is currently designed for a single user interacting with a single main agent.
An important future direction is extending this setting to support collaboration among multiple users, multiple agents, or both.
Such a setting raises new challenges in coordination, communication, provenance tracking, and conflict resolution, but it could also make agentic research workflows more effective for team-based projects.

\section{Related Work}
\label{sec:related_work}

We survey three bodies of work: AI systems that produce mathematical results autonomously (\Cref{sec:related_ai_math}), research on mathematicians actively using AI in their workflow (\Cref{sec:related_collab}), and agentic frameworks for scientific discovery (\Cref{sec:related_agentic}).
For broader surveys of AI for mathematics and scientific discovery, we refer to \citet{juAIMathematicsProgress2026}, \citet{carboneAdvancingMathematicsResearch2025}, and \citet{zhengAutomationAutonomySurvey2025}.

\subsection{AI Generating Mathematics}
\label{sec:related_ai_math}

\paragraph{Competition-level mathematics.}

In recent years, progress in AI mathematical reasoning has been especially visible in competition-level mathematics, where performance is relatively easy to compare because problems typically have a single, closed-form final answer that can be scored automatically.\footnote{Correct final answers need not imply correct proofs \citep{dekoninck2026}.}
Early results came from specialized systems: AlphaProof~\citep{hubertOlympiadlevelFormalMathematical2025} combined reinforcement learning with the Lean proof assistant to reach silver-medal performance at the 2024 IMO, while AlphaGeometry~\citep{trinhSolvingOlympiadGeometry2024a} and AlphaGeometry2~\citep{chervonyiGoldmedalistPerformanceSolving2025} paired a neural model with a symbolic deduction engine to achieve gold-medalist performance on historical olympiad geometry.
More recently, the emphasis has shifted toward off-the-shelf frontier models strengthened by verification and refinement: \citet{huang2025a} report a model-agnostic pipeline that, with Gemini~2.5~Pro, Grok-4, or GPT-5, solves five out of six problems on the 2025 IMO under contamination-avoiding protocols.
In parallel, proprietary systems such as Aristotle~\citep{achimAristotleIMOLevel2025} combine informal reasoning with formal verification to achieve gold-medal-equivalent performance on the 2025~IMO.
Finally, the same verification-first approach is now claimed at the undergraduate level: \citet{axiomprover2025putnam} reports that AxiomProver produced Lean-checked solutions to all Putnam~2025 problems (a perfect $120/120$).\footnote{cf. \url{https://axiommath.ai/territory/from-seeing-why-to-checking-everything}}
To move beyond competition-style evaluation, recent benchmarks increasingly probe research-level questions arising in active mathematical workflows, such as the encrypted, author-curated problem set in \emph{First Proof}~\citep{abouzaid2026}.

\paragraph{Constructions and algorithms.}
Beyond proving theorems, AI has generated novel mathematical constructions and faster classical algorithms by \emph{searching over programs}: an LLM proposes candidate code, an automated evaluator scores it, and an iterative loop improves the best candidates. FunSearch~\citep{romera-paredesMathematicalDiscoveriesProgram2024} introduced this template, yielding new constructions for the cap set problem and improved online bin packing heuristics. AlphaEvolve~\citep{novikovAlphaEvolveCodingAgent2025} scales the same evolutionary idea; in large-scale mathematical experiments it rediscovered best-known solutions across 67 problems and improved several, including autocorrelation inequalities~\citep{georgievMathematicalExplorationDiscovery2025}. 
Recent open-source works have proposed methodological extensions, including OpenEvolve, ShinkaEvolve, ThetaEvolve, DeltaEvolve, and AdaEvolve~\citep{openevolve,lange2025b,wang2025a,jiang2026a,cemri2026}.
Most such systems are \emph{closed-loop} and largely \emph{non-interactive}: progress comes from automated propose--evaluate iterations rather than back-and-forth dialogue with a human.
Related approaches have also produced faster algorithms: AlphaTensor~\citep{fawziDiscoveringFasterMatrix2022} discovered efficient tensor decompositions for matrix multiplication, and AlphaDev~\citep{mankowitzFasterSortingAlgorithms2023} found improved sorting routines now deployed in production software.

\paragraph{Data-driven and learning-augmented mathematics.}
A complementary line of work uses AI to generate candidate mathematical objects from data, whose correctness is then verified either automatically (via symbolic or optimization-based methods) or by human experts.
Examples include data-driven conjecturing and candidate filtering~\citep{daviesAdvancingMathematicsGuiding2021, mishraMathematicalConjectureGeneration2023, chuharski2024mining}, learning-augmented Lyapunov, Sum-of-Squares, and Border basis pipelines~\citep{alfaranoGlobalLyapunovFunctions2024, zouAnalyticalLyapunovFunction2025, pelleriti2025a, kera2025a}, neural-guided discovery of six-colorings for the Hadwiger--Nelson problem~\citep{mundinger2024extending, mundinger2025neural}, and ML+high-precision optimization uncovering unstable self-similar solutions in fluid dynamics~\citep{wangDiscoveryUnstableSingularities2025}.
Symbolic regression further extracts interpretable laws from data~\citep{udrescuAIFeynmanPhysicsinspired2020, ruanDiscoveringPhysicalLaws2026}.

\paragraph{Formal theorem proving and autoformalization.} 
A rich ecosystem of LLM-based formal proving tools has emerged around \emph{Lean 4}~\citep{lean4}.
LeanDojo~\citep{yangLeanDojoTheoremProving2023} provides an interface to Lean proof states and retrieval over mathlib~\citep{mathlib}, while Lean Copilot~\citep{songLeanCopilotLarge2025} integrates LLM assistance into the Lean workflow.
Dedicated provers include DeepSeek-Prover~\citep{xinDeepSeekProverAdvancingTheorem2024}, which leverages large-scale synthetic proof data, and DeepSeek-Prover-V2~\citep{ren2025}, which adds reinforcement learning with explicit subgoal decomposition and introduces ProverBench for evaluation. Goedel-Prover-V2~\citep{lin2025a} scales expert iteration with scaffolded data synthesis and verifier-guided self-correction.
Complementary directions focus on knowledge reuse and structured reasoning: LEGO-Prover~\citep{wangLEGOProverNeuralTheorem2023} builds and reuses a growing library of verified lemmas, while Hilbert~\citep{varamballyHilbertRecursivelyBuilding2025} connects informal reasoning with formal verification through recursive decomposition. TheoremLlama~\citep{wangTheoremLlamaTransformingGeneralPurpose2024} and Mathesis~\citep{xuejunMathesisFormalTheorem2025} explore adapting general-purpose models and end-to-end pipelines from natural language to Lean proofs.
Recent \emph{agentic} frameworks emphasize tool use and iterative compiler-feedback loops rather than one-shot generation: APOLLO~\citep{ospanov2025} performs modular proof repair and sub-lemma isolation, Ax-Prover~\citep{breen2025a} uses multi-agent tool-based proving across scientific domains, and LeanAgent~\citep{kumarappan2025} studies continual adaptation across evolving repositories. In a different direction, LeanProgress~\citep{george2026} guides search by predicting proof progress to improve performance on long proofs.
On the data side, MUSTARD~\citep{huangMUSTARDMasteringUniform2024} generates uniform theorem-and-proof training data with formal verification.
For evaluation, miniF2F~\citep{zheng2022} and PutnamBench~\citep{tsoukalas2024} provide competition-style benchmarks, while SorryDB introduces a dynamically updating stream of open \texttt{sorry} tasks mined from real-world Lean projects, mitigating contamination.
Autoformalization, i.e., translating informal mathematics into machine-checkable form, was shown to be feasible with LLMs by \citet{wuAutoformalizationLargeLanguage2022}. Recent work addresses this through dependency-graph decomposition~\citep{wangAriaAgentRetrieval2025}, chain-of-states proof translation~\citep{wangTranslatingInformalProofs2025}, and evaluation on real-world mathematical definitions~\citep{zhangAutoformalizationWildAssessing2025}. Agentic end-to-end pipelines such as MerLean~\citep{ren2026} extend this to scientific domains. We refer to \citet{wengAutoformalizationEraLarge2025} for a comprehensive survey.

\paragraph{Frontier systems and research-level evaluation suites.} 
Beyond competition benchmarks, several recent efforts target \emph{research-level} mathematics.
\emph{First Proof}~\citep{abouzaid2026} introduces an author-curated set of ten questions arising naturally in the authors' research, with answers not publicly released.
Other benchmarks include continuously refreshed collections drawn from arXiv papers (RealMath~\citep{zhang2025c}) and curated sets of exceptionally challenging, unpublished problems reviewed by domain experts (FrontierMath~\citep{glazer2025}).
Aletheia was evaluated directly on \emph{First Proof}: roughly three weeks after the challenge was introduced, \citet{feng2026a} report that Aletheia autonomously solved six out of ten problems.
Notably, some of these results are now accompanied by machine-checked proofs: for example, \citet{sothanaphan2026} provide a Lean formalization of a resolution of an Erd\H{o}s problem attributed to \citet{achimAristotleIMOLevel2025}.

\subsection{Mathematicians Using AI}
\label{sec:related_collab}

\paragraph{Frameworks and perspectives.}
The literature on AI and mathematical practice is broad, so we highlight only those lines of work most directly relevant to our framework.
\citet{haaseHumanAICoCreativityExploring2026} propose four levels of human-AI co-creativity: Digital Pen, AI Task Specialist, AI Assistant, and AI Co-Creator. These categories provide a conceptual vocabulary that we build on in \Cref{sec:levels}. Their treatment is intentionally broad and domain-agnostic, serving primarily as a conceptual template to which domain-specific details can be added. \citet{henkelMathematiciansAssistantIntegrating2025} offer a complementary perspective from mathematics, arguing that AI should augment rather than replace mathematical reasoning and proposing five guiding principles for its responsible use.
\citet{nooraniHumanAICollaborativeUncertainty2025} formalize the complementary strengths of humans and AI in uncertainty quantification, providing theoretical guarantees for collaborative prediction.
Most recently, \citet{avigadMathematiciansAgeAI} consider recent developments in AI-driven mathematics and argue that mathematicians should remain actively involved in the use of these systems.
Our work shares these perspectives but addresses a different question: given these emerging capabilities, how should a working researcher use them in practice?

\paragraph{Documented case studies.}
Over the past several months, a growing number of papers have documented how mathematicians interact with chat-based AI systems to obtain new research results \citep{bubeck2025a, diez2025a, alexeev2026, ivanisvili2025a, feldman2025a, salim2025a, dobribanSolvingResearchProblem2025, schmitt2025}.
More specialized agentic systems with varying degrees of autonomy are also being developed \citep{liuAIMathematicianPartner2025, fengAutonomousMathematicsResearch2026} and have already produced new mathematical results \citep{lee2026, feng2026}.
AI coding agents provide yet another pathway by enabling large computational searches: \citet{knuth2026claudescycles} report that Claude solved an open Hamiltonian cycle decomposition problem through iterative exploration. These examples likely represent only a small fraction of emerging workflows.

\subsection{Agentic Research Frameworks}
\label{sec:related_agentic}

\paragraph{Automated scientific discovery.}
\citet{luAIScientistFully2024} introduced \emph{The AI Scientist}, an end-to-end system that generates hypotheses, runs experiments, and writes papers; its successor~\citep{yamadaAIScientistv2WorkshopLevel2025} reported an AI-generated paper accepted at a peer-reviewed workshop.
Subsequent systems explore adjacent design points, from semi-automated, code-centric experimentation (CodeScientist~\citep{jansen2025}) to end-to-end agent pipelines that incorporate explicit mechanisms for human feedback and cumulative reporting \citep{schmidgall2025,schmidgall2025a}.
AlphaApollo~\citep{zhou2026a} combines multi-turn tool use, reinforcement learning, and iterative evolution with tool-assisted verification, showing improved performance on several mathematical reasoning benchmarks.
As these pipelines grow more complex, rigorous \emph{benchmarking} has emerged as a central challenge, with recent work proposing evaluations that target both full workflows and their individual steps~\citep{chenScienceAgentBenchRigorousAssessment2025,bragg2025}.
Taken together, these works highlight a common requirement: agent outputs must be checkable (e.g., as code, logs, or derived claims) and include explicit points for verification and human steering, rather than being treated as opaque end-to-end generations.
Karpathy's \emph{autoresearch} exemplifies a minimalist variant: an agent iteratively modifies a single file, runs fixed-budget training, and keeps or discards based on validation performance~\citep{karpathy_autoresearch_2026}.
Our framework targets the complementary regime of multi-file, multi-objective research with structured reporting and verification.
For broader context, we refer to recent surveys~\citep{ferrag2025,zheng-etal-2025-automation}.

\paragraph{Agentic coding tools.}

Terminal-based coding agents such as Claude Code, OpenCode, Codex CLI, and Gemini CLI~\citep{ClaudeCodeOverview,AnomalycoOpencode2026,CodexAICoding,BuildDebugDeployGeminiCLI} extend AI assistance beyond chat by enabling users~\citep{handa2025interviewer} (software engineers, analysts, and researchers alike) to delegate work within a persistent local project.
These agents can read and edit files and invoke development tools (e.g., shells, test runners, linters, and formatters) from within a CLI interface, producing inspectable artifacts such as patches, diffs, and test outputs.
This inspectable, file-based workflow is central to our setting: it enables reproducible iteration and makes it possible to attach verification hooks (tests, proofs, consistency checks) directly to the agent's actions.
A key recent development is the growth of long-running autonomy: in Claude Code, the 99.9th-percentile turn duration nearly doubled from under 25 to over 45 minutes between October 2025 and January 2026~\citep{anthropic2026agents}, reducing the need for constant supervision while increasing the importance of robust guardrails.
Finally, these tools separate the underlying model from a repository-scoped instruction file, allowing us to express our framework as a portable, model- and harness-agnostic procedure that applies across Claude Code, OpenCode, Codex CLI, and related CLI agents.

\section*{Acknowledgments}
The frameworks, approaches, and insights presented here have been developed in the context of the MATH+ project \emph{Agentic AI in Mathematics}.\footnote{\url{https://iol.zib.de/project/agentmath.html}} This research was partially supported by the Deutsche Forschungsgemeinschaft (DFG) through the DFG Cluster of Excellence MATH+ (EXC-2046/1, EXC-2046/2, project id 390685689), as well as by the German Federal Ministry of Research, Technology and Space (research campus Modal, fund number 05M14ZAM, 05M20ZBM) and the VDI/VDE Innovation + Technik GmbH (fund number 16IS23025B).

\bibliography{references}
\bibliographystyle{iclr2026_conference}

\end{document}